\documentclass[runningheads]{llncs}

\usepackage{eccv}

\usepackage{eccvabbrv}

\usepackage{graphicx}
\usepackage{booktabs}

\usepackage[accsupp]{axessibility}  

\usepackage{hyperref}

\usepackage{orcidlink}

\usepackage{latexsym}
\usepackage{kotex}
\usepackage{array}
\usepackage{makecell}
\usepackage{multicol}
\usepackage{nicefrac}  
\usepackage{mathtools}
\usepackage{amsfonts}
\usepackage{dsfont}
\usepackage{cuted}
\usepackage{capt-of}
\usepackage{xspace}
\usepackage{cases}
\usepackage{pifont}
\usepackage{kotex}
\usepackage{multirow}
\usepackage{breqn}
\usepackage{caption} 
\usepackage{color, colortbl}
\usepackage{enumitem}
\usepackage{mathtools}
\usepackage{url}
\usepackage{kotex}
\graphicspath{ {images/} }
\usepackage{multirow}
\usepackage{array}
\usepackage{boldline}
\usepackage{amsmath,amssymb}
\usepackage{algpseudocode}
\usepackage[utf8]{inputenc}
\usepackage[linesnumbered,ruled]{algorithm2e}
\SetKwRepeat{Do}{do}{while}

\definecolor{Gray}{gray}{0.9}
\definecolor{lightblue}{rgb}{0.87, 0.92, 0.97} 
\newcolumntype{g}{>{\columncolor{Gray}}c}

\newcommand{\cmark}{\ding{51}}
\newcommand{\xmark}{\ding{55}}

\usepackage{floatrow}

\usepackage{float}
\floatstyle{plaintop}
\restylefloat{table}
\def\Plus{\texttt{+}}
\def\Minus{\texttt{-}}

\usepackage{xspace}
\newcommand{\spinewebdata}{{AASCE}\xspace}
\newcommand{\buudataAP}{{BUU-AP}\xspace}
\newcommand{\buudataLA}{{BUU-LA}\xspace}

\newcommand{\baselineikename}{{Kim et al.}\xspace}

\newcommand{\iterone}{{\texttt{-i1}}\xspace}
\newcommand{\itertwo}{{\texttt{-i2}}\xspace}
\newcommand{\iterthree}{{\texttt{-i3}}\xspace}

\newcommand{\uczero}{{\texttt{-}}\xspace}
\newcommand{\ucone}{{\texttt{UC1}}\xspace}
\newcommand{\uctwo}{{\texttt{UC2}}\xspace}
\newcommand{\ucthree}{{\texttt{UC3}}\xspace}
\newcommand{\ucfour}{{\texttt{UC4}}\xspace}

\newcommand{\discriminator}{{detector}\xspace}
\newcommand{\corrector}{{corrector}\xspace}

\newcommand{\Discriminator}{{Detector}\xspace}
\newcommand{\Corrector}{{Corrector}\xspace}

\newcommand{\misvertex}{{vertex misidentification}\xspace}
\newcommand{\misbone}{{bone misidentification}\xspace}
\newcommand{\lrinv}{{left-right inversion}\xspace}

\newcommand{\KeyBot}{{KeyBot}\xspace}
\newcommand{\ReVert}{{KeyBot}\xspace}

\usepackage{xr}

\usepackage{tabularx}
\newcolumntype{Y}{>{\centering\arraybackslash}X}
\newcolumntype{L}{>{\raggedright\arraybackslash}X}
\newcolumntype{R}{>{\raggedleft\arraybackslash}X}

\begin{document}

\title{Bones Can't Be Triangles:\texorpdfstring{\\}{}Accurate and Efficient Vertebrae Keypoint Estimation through Collaborative Error Revision}

\titlerunning{Bones Can't Be Triangles}

\author{Jinhee Kim*\inst{1}\orcidlink{0000-0002-4495-0273} \and
Taesung Kim*\inst{1}\orcidlink{0000-0002-4976-6459} \and
Jaegul Choo\inst{1}\orcidlink{0000-0003-1071-4835}}
\authorrunning{J.~Kim et al.}
\institute{Korea Advanced Institute of Science and Technology\\
\email{\{seharanul17,~zkm1989,~jchoo\}@kaist.ac.kr}\\
* indicates equal contributions.
}

\maketitle
\begin{figure}[h!]
\vspace{-0.33in}
\centering
\includegraphics[width=1.0\linewidth]{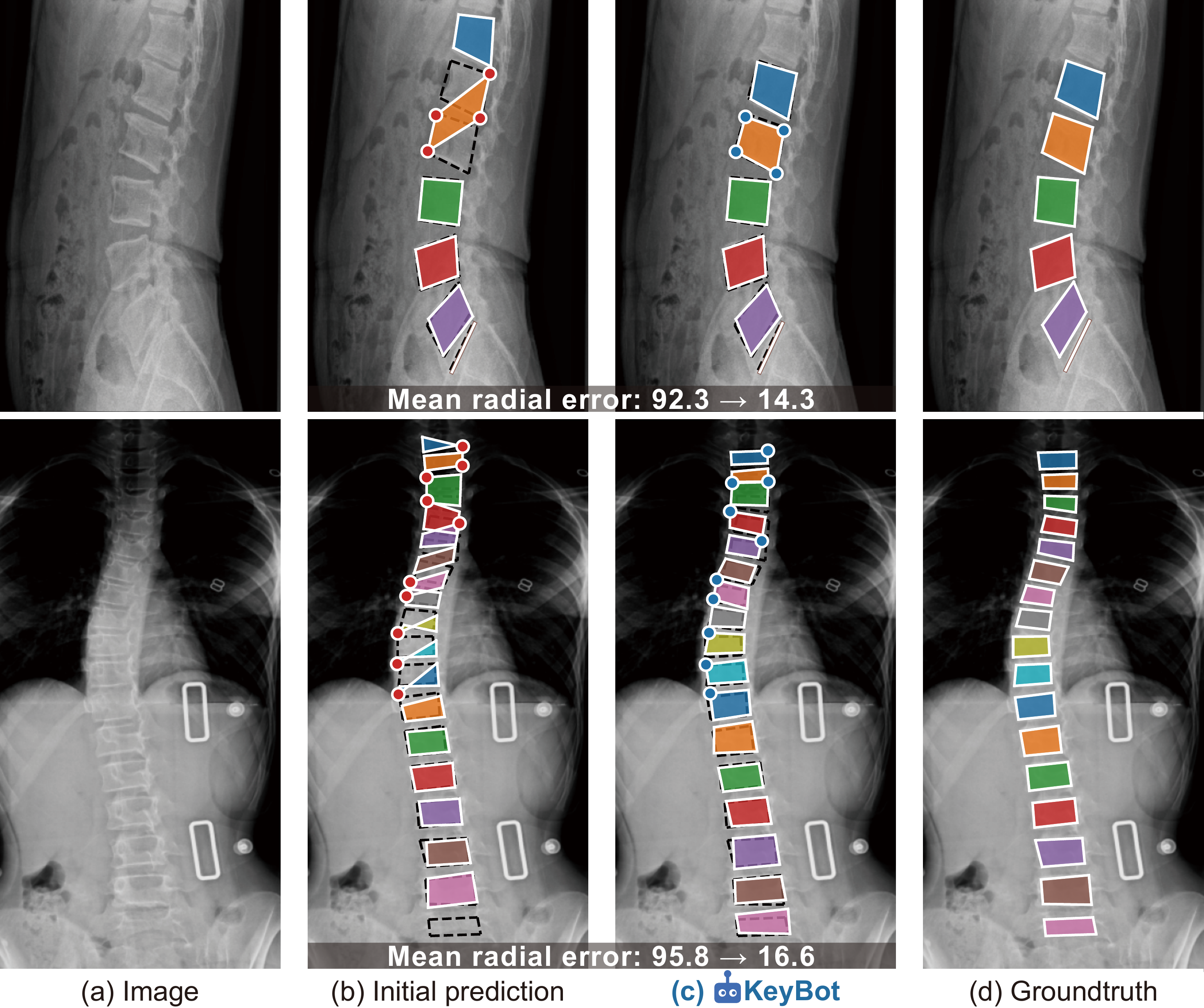}
    \caption{Vertebrae keypoint refinement results on BUU-LA~\cite{klinwichit2023buu} and AASCE~\cite{spinewebdataset}. Initial predictions exhibit significant errors in vertebra shapes due to difficulty identifying individual vertebrae. KeyBot effectively reduces these errors without user input.
    } 
    \label{fig:intro}
\end{figure}

\begin{abstract}
 \vspace{-0.35in}
Recent advances in interactive keypoint estimation methods have enhanced accuracy while minimizing user intervention. However, these methods require user input for error correction, which can be costly in vertebrae keypoint estimation where inaccurate keypoints are densely clustered or overlap.
We introduce a novel approach, KeyBot, specifically designed to identify and correct significant and typical errors in existing models, akin to user revision. By characterizing typical error types and using simulated errors for training, KeyBot effectively corrects these errors and significantly reduces user workload.
Comprehensive quantitative and qualitative evaluations on three public datasets confirm that KeyBot significantly outperforms existing methods, achieving state-of-the-art performance in interactive vertebrae keypoint estimation. The source code and demo video are available on our \href{https://ts-kim.github.io/KeyBot/}{\textbf{project page}}.
\end{abstract}

\section{Introduction}
\label{sec:intro}
Accurate vertebrae keypoint estimation from X-ray images is crucial for effective medical diagnosis and treatment planning~\cite{med_4,cvm2}, with errors having significant impacts on clinical decisions, such as in spinal surgeries and pain management~\cite{guo2021keypoint,liu2022multi,pisov2020keypoints}.
This task presents a distinct set of challenges due to the complex anatomy of the spine~\cite{zhang2024dual,rahmaniar2023auto}. 
The similar and repetitive shapes of vertebrae complicate distinguishing between individual vertebrae and accurately estimating their keypoints, as shown in Fig.~\ref{fig:intro}. This requires a deep understanding of sequential context and anatomy, unlike more distinct keypoints found in human pose~\cite{jin2020whole,guo2023back,shi2022end,posefix} or facial keypoint~\cite{colaco2022deep,albiero2021img2pose} estimation tasks.

\begin{figure}[t!]
\includegraphics[width=0.85\textwidth]{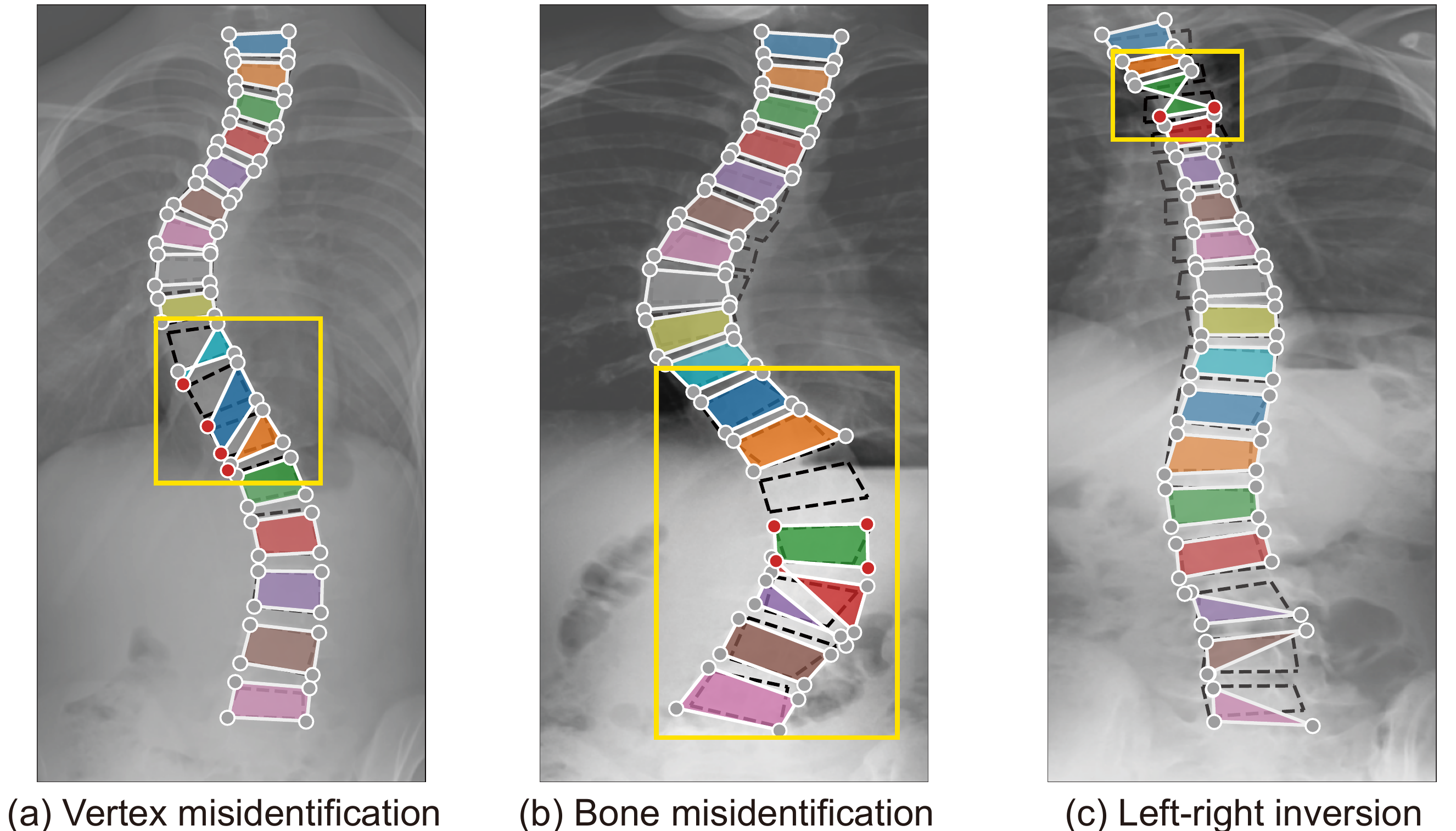}
\caption{Error types in vertebrae keypoint estimation, illustrated on the \spinewebdata dataset.} \label{fig:error_type}
\end{figure}

\begin{figure}[t!]
\includegraphics[width=1.0\textwidth]{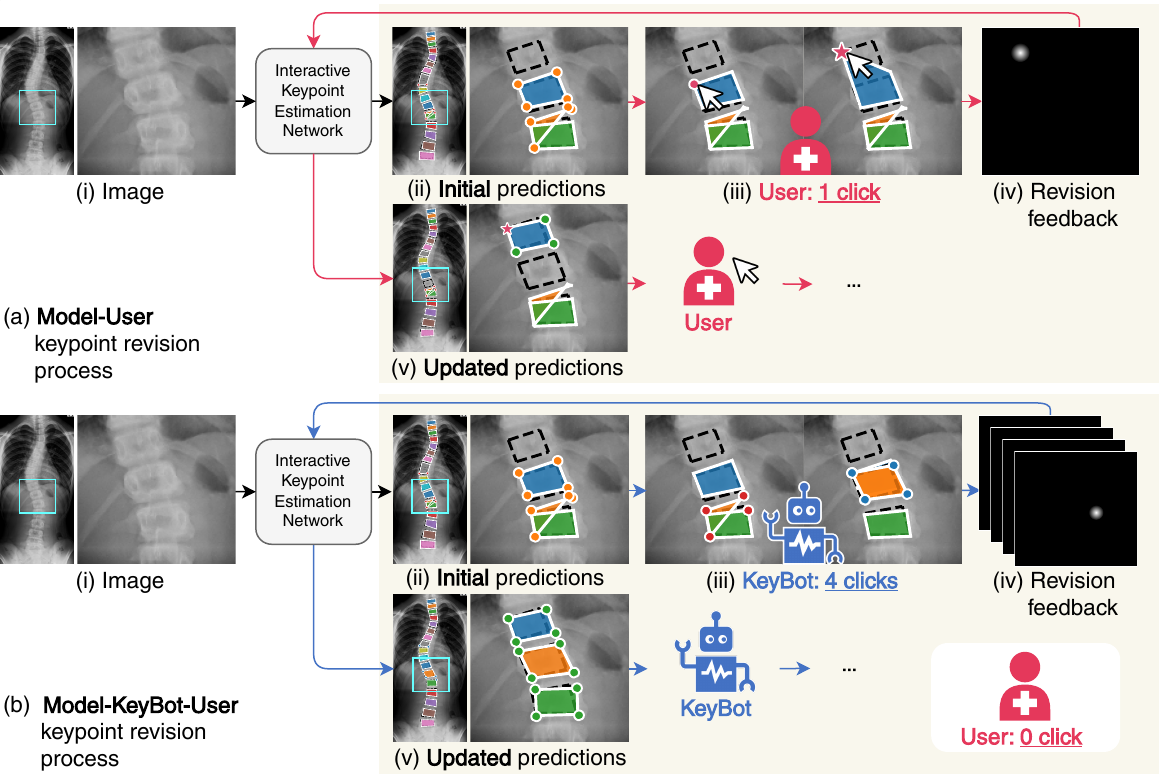}
\caption{Overview of (a) the interactive keypoint estimation and (b) the KeyBot framework. The interaction model generates initial keypoint predictions from an image, which are then revised by users, leading to updated results. This approach requires user input for error correction. 
In contrast, KeyBot offers a cost-efficient feedback mechanism via automated, rapid, and iterative refinement without user input. It independently identifies and corrects major errors, allowing users to focus on the final adjustments.
} \label{fig:flow}
\end{figure}

Such challenges often result in significant errors during automated vertebrae keypoint estimation, including vertex misidentification, bone misidentification, and left-right inversion, as illustrated in Fig.~\ref{fig:error_type}.   
Manual revision of these errors is time-consuming and labor-intensive, especially as the number of vertebrae increases. Missing even a single vertebra can lead to extensive adjustments, thereby significantly increasing the time and effort required for accurate annotation.
To mitigate these, interactive keypoint estimation approaches have been developed, significantly improving performance with minimal user input~\cite{kim2022morphology_mike,yang2023neural_clickpose}.
These methods aim to automatically refine inaccurate keypoints with minimal user feedback, using the feedback as a hint to adjust the remaining errors, as shown in Fig.~\ref{fig:flow}(a).
However, they heavily rely on users to identify and correct errors, which can be time-consuming and labor-intensive, especially when erroneous keypoints are densely clustered or overlapped. 

In response, we introduce a novel method named \textbf{KeyBot}, designed to autonomously detect and correct specific errors in vertebrae keypoint estimation before user evaluation, building upon existing interactive keypoint estimation models, as illustrated in Fig.~\ref{fig:flow}(b). 
Initially, the interaction model predicts keypoints based on the input image. During refinement, KeyBot autonomously pre-evaluates these predictions, identifies specific errors, and provides corrective feedback similar to user input. 
In contrast to previous models requiring at least one user input, KeyBot autonomously refines model predictions before user input.

We generate synthetic data characterizing typical error types, \misvertex, \misbone, and \lrinv, and train KeyBot using this data to focus specifically on predefined error types.  
Independently trained from the interaction model, KeyBot effectively identifies errors that might be overlooked by the interaction model, significantly reducing the user's revision efforts. 
Rather than correcting errors from scratch, users can focus on verifying and fine-tuning results, with KeyBot having already eliminated major errors.

KeyBot delivers quick, cost-effective, and iterative feedback without user input, which is crucial for complex cases requiring multiple refinements. This automated feedback is typically much faster than user review, significantly enhancing workflow efficiency.
The collaborative feedback loop between the user, KeyBot, and the interaction model enriches the process by combining expert medical insights with KeyBot's focus on typical error types. This synergy addresses a broad range of potential errors and enhancements, leading to an overall improvement in keypoint estimation accuracy.

Through extensive experiments on three public datasets~\cite{spinewebdataset,klinwichit2023buu}, KeyBot demonstrates significant improvements in vertebrae keypoint estimation accuracy. 
The model reduces the mean radial error (MRE) by 19\% and decreases the number of user clicks (NoC) required for target performance by 17\% on the \spinewebdata dataset, achieving state-of-the-art performance.  
These results highlight the significance of KeyBot in advancing vertebrae keypoint estimation, offering an advanced, user-centric solution to this unique and complex challenge.

In summary, our contributions are: (1) We introduce a novel method, KeyBot, designed to provide automated corrections for specific errors in vertebrae keypoint estimation. This enhances existing interactive keypoint estimation frameworks and improves overall keypoint estimation accuracy and efficiency. (2) We train KeyBot on synthetic data representing common error types in vertebrae, enabling it to effectively identify and address these errors. (3) KeyBot significantly reduces the need for user intervention during the keypoint annotation process.
(4) Extensive evaluation across three public datasets confirms the efficacy of our proposed method in improving keypoint estimation accuracy and significantly reducing user intervention.

\section{Related work}
\noindent \textbf{Automated error refinement.}
Despite the advancements in deep learning, errors in model predictions remain a formidable challenge.
To mitigate these inaccuracies, a specialized area of research dedicated to refining prediction results has emerged, leading to the development of advanced methods for identifying and correcting errors in model outputs~\cite{posefix,qu2024abdomenatlas,xu2019structured,daniele2023refining}.
For example, Qu et al.~\cite{qu2024abdomenatlas} introduced a method leveraging attention maps to detect label errors in CT data, markedly improving annotation efficiency and facilitating the collection of extensive datasets.
Similarly, efforts to refine keypoint estimation have been noteworthy.
For instance, Ronchi et al.~\cite{ruggero2017benchmarking} analyzed human pose estimation results to categorize three main types of errors, while Moon et al.~\cite{posefix} developed a method that specifically addresses and corrects these categorized errors, achieving state-of-the-art human pose estimation performances.

Despite these advancements, estimating vertebrae keypoints from X-ray images introduces additional challenges due to the complex sequential structure of the spine. Error patterns, such as the mislocalization of an entire vertebra, often require extensive user inputs for correction.
In response, we propose KeyBot, a method tailored for revising errors specific to vertebrae keypoint estimation, effectively addressing the unique challenges of this task.

\noindent \textbf{User interaction.}
To maximize the accuracy of predictions while minimizing the need for user input, the development of interaction-based models is paramount. 
Interactive segmentation is a particularly active area within this domain. 
Backpropagating refinement schemes such as BRS~\cite{brs} and f-BRS~\cite{fbrs} have improved interactive image segmentation by incorporating user hints through backpropagation. Similarly, RITM~\cite{ritm} has shown superior performance by integrating an iterative refinement process into its training process. 
In the field of keypoint estimation, interactive keypoint estimation methods have been researched to reduce user input through interactive refinement frameworks.
Kim et al.~\cite{kim2022morphology_mike} pioneered an interactive framework for keypoint estimation, significantly improving the accuracy and efficiency of the keypoint annotation process in medical datasets. ClickPose~\cite{yang2023neural_clickpose} developed an interactive keypoint estimation model for multi-person 2D keypoints, significantly reducing annotation costs. 

Recently, Liu et al.~\cite{liu2022pseudoclick} introduced a method that generates pseudo clicks to mimic human interactions for refining segmentation predictions autonomously, thereby minimizing the required number of user clicks.
However, their error decoder is trained end-to-end and can thus be affected by inherent biases or limitations of the segmentation module.
To address this challenge, our KeyBot is trained independently to correct simulated errors, which allows KeyBot to provide corrective feedback free from any potential biases present in the keypoint estimation model. 
Unlike traditional interactive keypoint estimation models that rely on users to identify and correct errors, KeyBot autonomously generates corrective feedback. This significant innovation not only enhances the keypoint estimation accuracy but also remarkably reduces the need for user interaction, thereby streamlining the error correction process.

\begin{figure}[t!]
\includegraphics[width=1.0\textwidth]{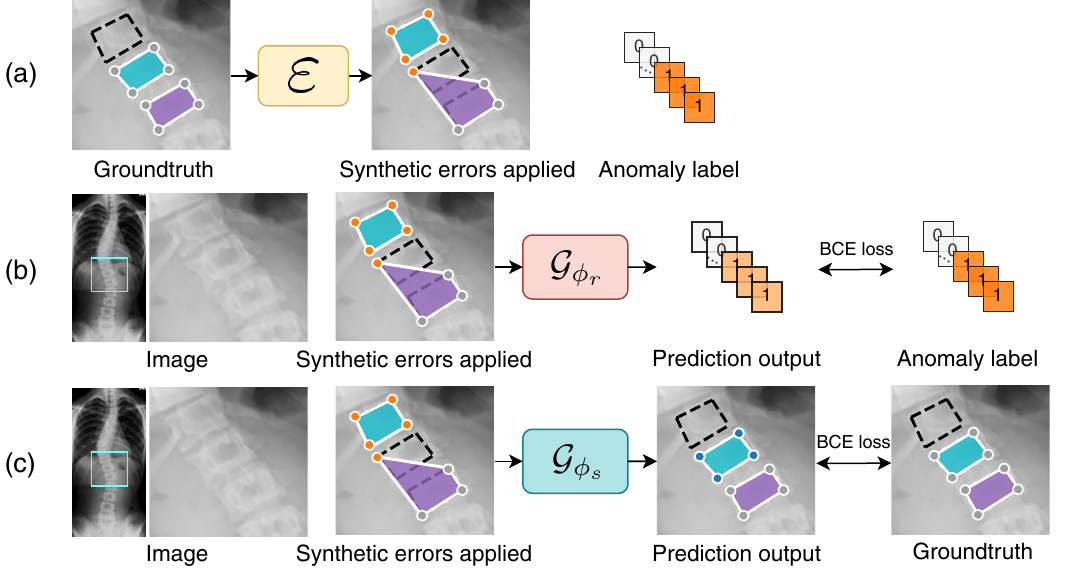}
\caption{Training overview of KeyBot. KeyBot consists of two main components, the detector and the corrector. (a) Synthetic errors are introduced to the groundtruth keypoints, creating inaccurate keypoints and corresponding anomaly labels. (b) The detector is trained to discern whether each input keypoint is accurate or not. (c) The corrector is trained to refine these inaccurate keypoints accurately.} \label{fig:method}
\end{figure}

\section{KeyBot}
We introduce KeyBot, a novel method designed to enhance vertebrae keypoint estimation by identifying and correcting errors prior to user intervention. We employ an existing interactive model that can incorporate user feedback for interactive modifications. Given model predictions, KeyBot identifies erroneous keypoints and generates their pseudo-corrections, which are then fed to the interaction model for updates. KeyBot consists of two components: The detector discerns whether each input keypoint is accurate, and the corrector refines input keypoints accurately. Both components are trained with synthetic errors to focus specifically on the target error types.

\subsection{Overview of the propose approach}\label{sec:overview}
\noindent \textbf{Interactive keypoint estimation framework.}  
Consider an interactive keypoint estimation model $\mathcal{F}_\theta$ with parameters $\theta$, operating on an image $\boldsymbol{x}$. The model aims to precisely estimate $K$ target keypoints through user-interactive refinement.
In each iteration $t$, the model generates the $t$-th prediction, $\boldsymbol{y}_t$, using three inputs: the image $\boldsymbol{x}$, user revisions $\boldsymbol{u}_{t}$, and its previous predictions $\boldsymbol{e}_{t}$: 
\begin{equation}
{\boldsymbol{y}_{t}} = \mathcal{F}_\theta(\boldsymbol{x}, {\boldsymbol{u}_{t}}, {\boldsymbol{e}_{t}}),
\end{equation}
where ${\boldsymbol{u}_{0}}\coloneqq 0$ and $\boldsymbol{e}_{0}\coloneqq 0$.
Subsequently, a user reviews the results, and the adjustment made to a keypoint position is encoded as Gaussian heatmaps $\boldsymbol{u}_{t}^i$, effectively capturing and representing the refined locations:
\begin{equation}
\boldsymbol{\rho}_{t\Plus1}, \boldsymbol{u}_{t\Plus1} =  \Psi(\boldsymbol{x},\boldsymbol{y}_{t},\boldsymbol{u}_{t}),
\label{eq:user}
\end{equation}
where the set $\boldsymbol{\rho}_{t}$ represents keypoint indices revised by the user over $t$ iterations.
Next, we selectively gather the \textit{known-to-be-false} predictions from $\boldsymbol{y}$ to $\boldsymbol{e}$ and reintroduce them as inputs in the next step to differentiate mispredictions~\cite{kim2022morphology_mike}:
\begin{equation}
\boldsymbol{e}_{t\Plus1}^i = \boldsymbol{y}_{t}^i \   \text{for} \  i \in \boldsymbol{\rho}_{t}. \label{eq:prev_everyupdate}
\end{equation}
Here, $\boldsymbol{e}$ denotes predictions modified by the user.
 These updates on $\boldsymbol{u}_{t}$ and $\boldsymbol{e}_{t}$ are critical for refining the model predictions, leading to the updated keypoint predictions reflecting the user feedback:  
\begin{equation}
{\boldsymbol{y}_{t\Plus1}} = \mathcal{F}_\theta(\boldsymbol{x}, {\boldsymbol{u}_{t\Plus1}}, {\boldsymbol{e}_{t\Plus1}}).
\end{equation}
This process repeats until sufficient accuracy is achieved or the predefined maximum user interaction $T$ is reached.

\noindent \textbf{The \KeyBot framework.}
KeyBot offers an automated keypoint refinement step that precedes and complements the existing user-based refinement, effectively reducing the need for user interaction in the process.
KeyBot consists of the detector $\mathcal{G}_{\phi_r}$ and corrector $\mathcal{G}_{\phi_s}$, which together identifies and corrects inaccurate keypoints predicted by $\mathcal{F}_\theta$.
Following each prediction iteration of $\mathcal{F}_\theta$, \KeyBot evaluates $\boldsymbol{y}_t$ and suggests necessary corrections, identifying erroneous keypoints $\boldsymbol{\nu}_{t,n\Plus1}$ and their pseudo corrections $\boldsymbol{z}_{t,n\Plus1}$: 
\begin{equation}
{\boldsymbol{\nu}_{t,n\Plus1}} = \mathcal{G}_{\phi_r}(\boldsymbol{x}, {\boldsymbol{y}_{t,n}}) \ \text{and}
 \ {\boldsymbol{z}_{t,n\Plus1}} = \mathcal{G}_{\phi_s}(\boldsymbol{x}, {\boldsymbol{y}_{t,n}}),
\end{equation}
with $\boldsymbol{y}_{t,0} \coloneqq \boldsymbol{y}_t$. This step is similar to the user interaction step outlined in Eq.~(\ref{eq:user}).
Here, \KeyBot can detect multiple inaccurate keypoints at once and simultaneously offer corrective feedback for each identified error.
If no inaccuracies are detected by the detector, \KeyBot concludes and $\boldsymbol{y}_{t,n}$ becomes the final prediction.
Revision information, $\boldsymbol{c}_{t,n}$, now includes both user ($\boldsymbol{u}_{t,n}$) and model ($\boldsymbol{z}_{t,n}$) corrections. Here, user modifications are prioritized as the definitive ground truth. Thus, keypoints already revised by the user are exempt from further updates by the pseudo-corrections from \KeyBot.
\begin{equation}
\boldsymbol{c}_{t,n\Plus1}^i = {\boldsymbol{z}_{t,n\Plus1}^i\   \text{for} \  i \in {\boldsymbol{\nu}_{t,n\Plus1}} \setminus\ \boldsymbol{\rho}_{t}}.
\end{equation} 
Previous work~\cite{ritm, kim2022morphology_mike} relies solely on the prediction at the most recent step as the previous prediction, as described in Eq.~(\ref{eq:prev_everyupdate}). However, this approach can result in losing track of erroneous predictions once they are corrected in the model. Hence, our approach accumulates predictions identified as erroneous by either KeyBot or the user across all iterations. These false predictions retain the history of incorrect predictions and serve to inform the model about specific prediction errors to avoid, thereby guiding it toward more accurate future predictions:
\begin{equation}
\boldsymbol{e}_{t,n\Plus1}^i = {\boldsymbol{y}_{t,n}^i\   \text{for} \ i \in {\boldsymbol{\nu}_{t,n\Plus1}} \setminus {\boldsymbol{\nu}_{t,:n}}  \setminus \boldsymbol{\rho}_{t}}.
\end{equation}
Here, $\boldsymbol{e}$ denotes predictions modified by either the user or KeyBot.
Finally, these are then fed back into the backbone interaction network for a new, corrected prediction akin to user feedback:
\begin{equation}
    {\boldsymbol{y}_{t,n\Plus1}} = \mathcal{F}_\theta(\boldsymbol{x}, {\boldsymbol{c}_{t,n\Plus1}}, {\boldsymbol{e}_{t,n\Plus1}}).
\end{equation}
Each iteration of KeyBot contributes progressively towards enhancing the overall accuracy and reliability of the keypoint predictions.
This cycle repeats until the detector detects no errors or the maximum KeyBot iteration count $N$ is reached. 
Upon the KeyBot process concludes, say, after $n\Plus1$ iterations, the prediction ${\boldsymbol{y}_{t,n\Plus1}}$ is established as the final output for the $t$-th iteration of the interactive keypoint estimation process. 
For the subsequent $(t\Plus1)$-th user-side phase, given new user feedback as Eq.~(\ref{eq:user}),
$\boldsymbol{c}_{t,n\Plus1}$ and $\boldsymbol{e}_{t,n\Plus1}$ are updated to $\boldsymbol{c}_{t\Plus1,0}$ and $\boldsymbol{e}_{t\Plus1,0}$, 
\begin{equation}
\boldsymbol{c}_{t\Plus1,0} = {\boldsymbol{u}_{t\Plus1}} \ \text{and} \  
\boldsymbol{e}_{t\Plus1,0}^i = {\boldsymbol{y}_{t,n\Plus1}^i \  \text{for} \ i \in \boldsymbol{\rho}_{t\Plus1}},
\end{equation}
and the keypoints are revised accordingly:
\begin{equation}{\boldsymbol{y}_{t\Plus1,0}} = \mathcal{F}_\theta(\boldsymbol{x}, {\boldsymbol{c}_{t\Plus1,0}}, {\boldsymbol{e}_{t\Plus1,0}}).
\end{equation}
The complete procedure is summarized in Algorithm~1 in Appendix~\ref{appendix:moremodeldetails}.

\subsection{Model design of \KeyBot}\label{sec:modeldesign}
KeyBot integrates the detector and corrector to identify potential errors and generates corrective feedback for their revision.

\noindent \textbf{Error detector.}
The error detector specializes in identifying up to $K$ erroneous keypoints, focusing on prevalent error types in vertebrae keypoint estimation. 
Our empirical analysis indicates that evaluating the entire bone structure at once is less effective for error detection. 
Therefore, the detector evaluates keypoints in smaller groups, examining $k \leq K$ keypoints simultaneously. 
This approach simplifies the task and improves its performance. 
During training, we randomly select a set of consecutive groundtruth keypoints $\boldsymbol{y}^*$ and either maintain their accuracy or introduce errors.
The cropped image area around these keypoints, along with their corresponding Gaussian keypoint heatmaps, are input to the detector. It then predicts the likelihood of error presence in each keypoint.

Let $\varepsilon$ denote the function that induces synthetic errors in groundtruth keypoints, and $\textit{v}$ represents the anomaly labels for the keypoints.
The detector performs multi-label binary classification to differentiate between normal and abnormal keypoints using a sigmoid function in its output layer. 
It is supervised to minimize the Binary Cross-Entropy (BCE) loss between the anomaly labels $\textit{v}$ and the predictions:
\begin{equation}
\phi_r \leftarrow \phi_r - \eta_r\nabla_{\phi_r}\mathcal{L}_{bce}(\textit{v}, \mathcal{G}_{\phi_r}(\boldsymbol{x}, \varepsilon(\boldsymbol{y}^*)),
\end{equation}
with $\eta_r$ representing the learning rate for the detector.

\begin{figure}[t!]
\centering
\includegraphics[width=1.0\textwidth]{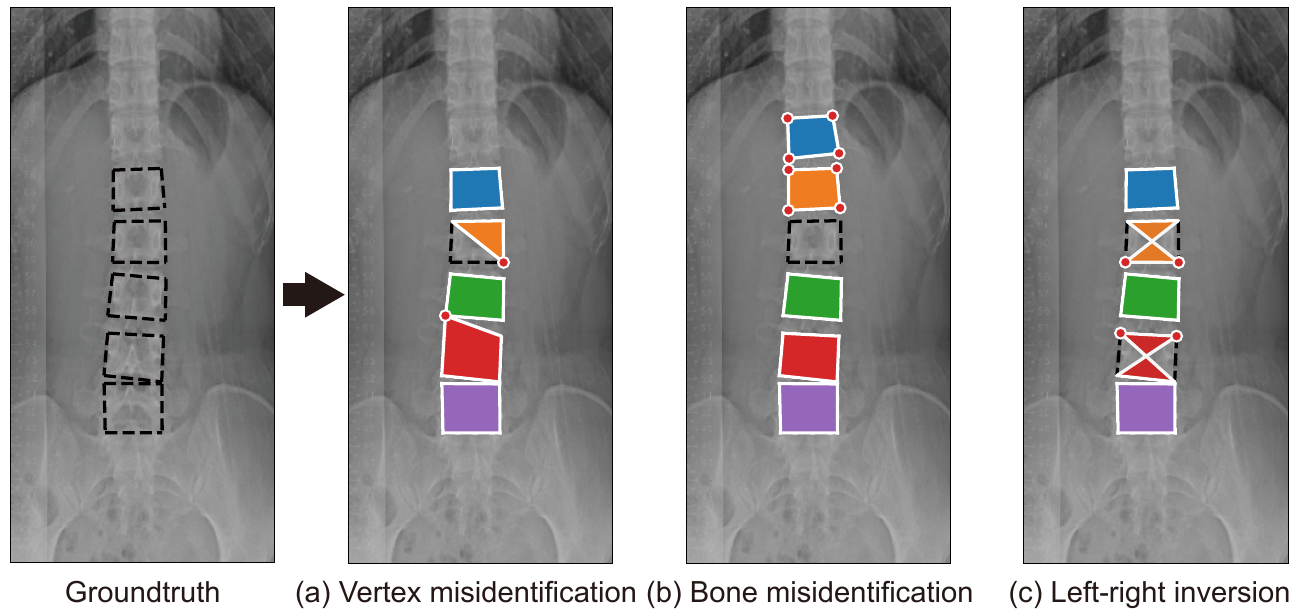}
\caption{Implementing realistic estimation inaccuracy simulations for training KeyBot.} \label{fig:negative_sample}
\end{figure}

During inference, the detector iteratively detects errors across all keypoints, examining $k$ keypoints at a time with a stride of $s$. To facilitate a comprehensive assessment of each keypoint in different contexts, we set the stride $s$ to be smaller than $k$. This results in multiple predictions for a single keypoint, and if any prediction indicates an anomaly, the keypoint is flagged as erroneous.

\noindent \textbf{Error \corrector.}
The error corrector aims to generate pseudo-corrections for keypoints inaccurately identified by the detector, thus providing feedback to the model $\mathcal{F}_\theta$. 
During training, we randomly transform the groundtruth keypoints to erroneous ones using $\varepsilon$. Taking these keypoints and the image $\boldsymbol{x}$ as inputs, the corrector is trained to revert these to their original, accurate state, focusing particularly on the typical error types implemented by $\varepsilon$. 
It generates $K$-channel heatmaps indicating reconstructed keypoint locations, employing a sigmoid function in its output.
Its learning objective is the BCE loss between the ground truth keypoint heatmaps and the reconstructed outputs:
\begin{equation}
\phi_s \leftarrow \phi_s - \eta_s\nabla_{\phi_s}\mathcal{L}_{bce}\big(\boldsymbol{y}, \mathcal{G}_{\phi_s}(\boldsymbol{x},  \varepsilon(\boldsymbol{y}^*)\big),
\end{equation}
where $\eta_s$ is the learning rate for the \corrector. 

During inference, the corrector refines predictions from the backbone model $\mathcal{F}_\theta$, and the output corresponding to inaccurately assessed keypoints by the detector is fed to $\mathcal{F}_\theta$ to serve as pseudo-corrections.

\subsection{Training KeyBot via realistic error simulation}\label{sec:errorsimulation}
KeyBot aims to address significant and typical errors in vertebrae keypoint estimation. 
The key idea is to train KeyBot to identify and correct three error types commonly observed in vertebrae keypoint estimation: \misvertex, \misbone, and \lrinv.   
During training, KeyBot is provided with both accurate keypoints and keypoints with synthetic errors. We randomly simulate the predefined error types, as shown in Fig.~\ref{fig:negative_sample}.
Rather than relying on highly variable and not clearly defined real errors, we use characterized and consistent synthetic errors. 

By employing this training approach, we expose KeyBot to a wide range of error scenarios, equipping it with the proficiency to detect and correct these errors effectively.
As a result, KeyBot provides a targeted evaluation of the keypoints predicted by the interaction model, focusing on specific error types.

\noindent \textbf{Vertex misidentification (misvertex).} Misvertex is a type of error where a keypoint is correctly identified on a vertex of a vertebrae but is incorrectly attributed to the wrong vertex. 
For instance, a keypoint intended for the first vertex is inaccurately predicted at the third vertex. 
To simulate misvertex, we randomly displace a keypoint to its neighbors within a $[\Minus r, \Plus r]$ index range for a predefined number $r$, excluding its original position. Keypoints moving beyond their index range wrap around cyclically.  

\noindent \textbf{Bone Misidentification (misbone).} Misbone errors represent errors where the entire vertebra is incorrectly classified.
Our simulation randomly shifts keypoints up or down by one vertebra.
This is achieved by selecting start and end keypoint indices and moving all-encompassed keypoints to an adjacent vertebra.

\noindent \textbf{Left-right inversion (lr-inversion).} LR-inversion occurs when keypoint pairs are incorrectly positioned on the opposite lateral side. To simulate LR-inversion in our model, we randomly interchange the positions of keypoint pairs that delineate one side of each vertebra.

\section{Experiments}

\begin{table*}[t!]
\caption{Performance comparison of mean radial error in keypoint estimation across three datasets. \texttt{UC} denotes the count of user clicks provided to the model.}
\label{table:mre_table}
\begin{center}
\resizebox{0.95\textwidth}{!}{%
\begin{tabular}{lccccc|ccc|ccc}
\toprule
\multicolumn{1}{l}{\multirow{2.3}{*}{Method}} & 
\multicolumn{5}{c}{\spinewebdata} &
\multicolumn{3}{c}{\buudataAP} &
\multicolumn{3}{c}{\buudataLA} \\
\cmidrule(l{2pt}r{2pt}){2-6} \cmidrule(l{2pt}r{2pt}){7-9}\cmidrule(l{2pt}r{2pt}){10-12}
&
\makecell{\uczero} &\makecell{\ucone} &\makecell{\uctwo} &\makecell{\ucthree} &\makecell{\ucfour} &
\makecell{\uczero} &\makecell{\ucone} &\makecell{\uctwo} &
\makecell{\uczero} &\makecell{\ucone} &\makecell{\uctwo} \\
\midrule\midrule
\multicolumn{1}{l}{$\text{BRS}$~\cite{brs}} 
    & 45.65 & 35.88 & 29.93 & 25.40 & 21.97
    & 51.22 & 30.91 & 36.17
    & 40.20 & 43.19 & 35.07
\\
\multicolumn{1}{l}{$\text{f-BRS}$~\cite{fbrs}}
    & 64.06 & 57.30 & 52.53 & 47.83 & 43.79
    & 44.05 & 24.99 & 17.52
    & 36.03 & 27.93 & 19.29 
\\
\multicolumn{1}{l}{RITM~\cite{ritm}} 
    & 56.03 & 37.96 & 31.03 & 26.76 & 23.79
    & 36.43 & 25.76 & 17.96
    & 23.27 & 17.79 &  9.89 
\\
\midrule  
\multicolumn{1}{l}{{\baselineikename~\cite{kim2022morphology_mike}}}
    & 51.58 & 30.60 & 25.78 & 21.60 & 19.08
    & 42.31 & 23.26 & 16.46
    & 23.43 & 14.29 & 10.29
\\
\multicolumn{1}{l}{{Click-Pose~\cite{yang2023neural_clickpose}}} 
    & 54.65 & 46.50 & 44.08 & 41.73 & 40.04
    & 32.72 & 29.30 & 26.20    
    & 33.70 & 21.62 & 17.38
\\
\midrule  
\rowcolor{lightblue}\multicolumn{1}{l}{{\ReVert\iterone}}
    & 44.18 & \textbf{25.93} & \textbf{20.66} & \textbf{18.03} & 16.37
    & 32.01 & 21.79 & 15.97
    & 18.77 & 13.39 &  9.12
\\
\rowcolor{lightblue}\multicolumn{1}{l}{{\ReVert\itertwo}}
    & 42.52 & 27.03 & 22.95 & 18.72 & \textbf{16.23}
    & 31.88 & \textbf{20.65} & \textbf{15.81}
    & 19.11 & 13.47 &  9.03
\\
\rowcolor{lightblue}\multicolumn{1}{l}{{\ReVert\iterthree}}
    & \textbf{41.70} & 26.59 & 25.02 & 21.23 & 16.76
    & \textbf{31.87} & 20.66 & 15.84
    & \textbf{18.74} & \textbf{13.36} &  \textbf{8.97}
\\
\bottomrule
\end{tabular}
}
\end{center}
\end{table*}

\subsection{Experimental Setup}\label{section:experimental_setup}

\noindent \textbf{Datasets.}\label{sec:dataset}
We rigorously validate our method across three publicly available real-world datasets.
The \textbf{\spinewebdata} dataset~\cite{spinewebdataset} consists of 608 spinal anterior-posterior radiographs, each annotated with 68 keypoints, marking the vertices of 17 vertebrae.
The BUU dataset~\cite{klinwichit2023buu} features two distinct subsets from 400 unique patients: \textbf{\buudataAP} with 400 anterior-posterior (AP) view radiographs, each annotated with 20 keypoints, and \textbf{\buudataLA} with 400 left lateral (LA) view radiographs, each annotated with 22 spinal keypoints. 
Images that exhibit critical keypoint annotation errors, such as those with incorrectly ordered keypoint indices similar to typical errors in keypoint estimation models, are excluded from the datasets.
Comprehensive details about the dataset, including the data split and examples of the excluded images, are available in Appendix~\ref{appen_sec:dataset}.

\noindent \textbf{Metrics.}
Following the evaluation protocol of previous work~\cite{kim2022morphology_mike, yang2023neural_clickpose}, we evaluate the model performance using the mean radial error (MRE) and the number of clicks (NoC). MRE computes the mean Euclidean distance between the groundtruth and predicted keypoints.
$\text{NoC}_{a}@b$ measures the average number of user clicks required to achieve a target MRE of $b$, given a maximum of $a$ clicks.

\noindent \textbf{Baseline models.}
We compare our method with state-of-the-art interactive keypoint estimation models, including the model proposed by Kim et al.~\cite{kim2022morphology_mike} and Click-Pose~\cite{yang2023neural_clickpose}. Additionally, we use interactive image segmentation models such as BRS~\cite{brs}, f-BRS~\cite{fbrs}, and RITM~\cite{ritm} as baselines. 

\noindent \textbf{Implementation details.} 
We utilize ResNet~\cite{resnet}-18 and DeepLab-V3~\cite{chen2017rethinking_deeplabv3} as the backbone for the detector and corrector, respectively.
We build on the pretrained model of Kim et al.~\cite{kim2022morphology_mike}, which is designed to estimate keypoints based on user feedback. All of the models are trained independently.
We apply the misvertex error type to train the detector and all three error types for the corrector. 
KeyBot\iterone, KeyBot\itertwo, and KeyBot\iterthree refer to KeyBot with one, two, and three iterations, respectively. Unless explicitly stated, the experimental results are conducted without user interaction. 
The training and inference are conducted on a single NVIDIA GeForce RTX 3090 GPU with 24GB memory.
More details on experimental settings, including training details, are provided in Appendix~\ref{appendix:moreexpdetails}.

\begin{table*}[t!]
\caption{{Comparison of required number of user clicks to reach target performance on the AASCE, BUU-AP, and
BUU-LA datasets.}}
\label{table:ikepublic}
\begin{center}
\resizebox{0.95\textwidth}{!}{%
\begin{tabular}{lcccc|cccc|cccc}
\toprule
\multicolumn{1}{l}{\multirow{3.3}{*}{Method}} & 
\multicolumn{4}{c}{\spinewebdata} &
\multicolumn{4}{c}{\buudataAP} &
\multicolumn{4}{c}{\buudataLA} \\
\cmidrule(l{2pt}r{2pt}){2-5} \cmidrule(l{2pt}r{2pt}){6-9}\cmidrule(l{2pt}r{2pt}){10-13}
\multicolumn{1}{c}{}&
\makecell{$\text{NoC}_{10}$\\@20} &
\makecell{$\text{NoC}_{10}$\\@30} & 
\makecell{$\text{NoC}_{10}$\\@40} &
\makecell{$\text{NoC}_{10}$\\@50} &
\makecell{$\text{NoC}_5$\\@15} &
\makecell{$\text{NoC}_5$\\@20} & 
\makecell{$\text{NoC}_5$\\@25} &
\makecell{$\text{NoC}_5$\\@30} &
\makecell{$\text{NoC}_5$\\@15} &
\makecell{$\text{NoC}_5$\\@20} & 
\makecell{$\text{NoC}_5$\\@25} &
\makecell{$\text{NoC}_5$\\@30} \\ 
\midrule\midrule
\multicolumn{1}{l}{$\text{BRS}$~\cite{brs}} 
    &  2.53 &  1.97 &  1.51 &  1.23
    &  2.80 &  1.51 &  0.92 &  0.61
    &  3.89 &  2.26 &  1.34 &  0.86
\\
\multicolumn{1}{l}{$\text{f-BRS}$~\cite{fbrs}} 
    &  7.16 &  5.30 &  4.21 &  3.48
    &  2.23 &  1.18 &  0.68 &  0.48
    &  1.24 &  0.61 &  0.44 &  0.40
\\
\multicolumn{1}{l}{RITM~\cite{ritm}} 
    &  3.20 &  2.37 &  1.87 &  1.38
    &  1.05 &  0.58 &  0.42 &  0.30
    &  0.57 &  0.24 &  0.17 &  0.14
\\
\midrule  
\multicolumn{1}{l}{{\baselineikename~\cite{kim2022morphology_mike}}}
    &  2.10 &  1.54 &  1.23 &  1.03
    &  0.94 &  0.54 &  0.37 &  0.32
    &  0.38 &  0.15 &  0.14 &  0.12
\\
\multicolumn{1}{l}{{Click-Pose~\cite{yang2023neural_clickpose}}} 
    &  4.15 &  3.56 &  3.20 &  2.97
    &  1.20 &  0.89 &  0.76 &  0.71
    &  1.45 &  0.70 &  0.56 &  0.45
\\
\midrule  
\rowcolor{lightblue}\multicolumn{1}{l}{{\ReVert\iterone}}
    &  1.84 &  1.26 &  0.92 &  \textbf{0.53}
    &  \textbf{0.87} &  0.51 &  0.33 &  0.28
    &  \textbf{0.34} &  \textbf{0.10} &  \textbf{0.06} &  \textbf{0.05}
\\
\rowcolor{lightblue}\multicolumn{1}{l}{{\ReVert\itertwo}}
    &  1.78 &  \textbf{1.21} &  \textbf{0.91} &  0.61
    &  \textbf{0.87} &  \textbf{0.48} &  \textbf{0.30} &  \textbf{0.24}
    &  0.35 &  0.11 &  \textbf{0.06} &  \textbf{0.05}
\\
\rowcolor{lightblue}\multicolumn{1}{l}{{\ReVert\iterthree}}
    &  \textbf{1.74} &  1.32 &  0.94 &  0.68
    &  \textbf{0.87} &  0.49 &  0.32 &  \textbf{0.24}
    &  \textbf{0.34} &  \textbf{0.10} &  \textbf{0.06} &  \textbf{0.05}
\\
\bottomrule
\end{tabular}
}
\end{center}
\end{table*}
\subsection{Experimental results and analysis}

\noindent \textbf{Comparison with state-of-the-art methods.}
In this section, we present a comparative analysis of interactive keypoint estimation performance between KeyBot and established baseline models by increasing the maximum KeyBot iteration count $N$ from one to three.
As shown in Table~\ref{table:mre_table}, KeyBot achieves state-of-the-art performance in mean radial error (MRE) across the AASCE, BUU-AP, and BUU-LA datasets. 
KeyBot consistently outperforms baseline models both in initial predictions and during subsequent user interactions. 
Remarkably, KeyBot with a single user click (\ucone) shows superior performance compared to BRS, f-BRS, RITM, and Click-Pose with two user clicks (\uctwo) on the AASCE dataset.
Furthermore, the performance of KeyBot improves with additional iterations, demonstrating its ability to refine predictions through multiple correction cycles.
These results highlight the effectiveness of KeyBot in enhancing keypoint estimation accuracy and reducing user intervention.

In terms of the number of clicks (NoC), as shown in Table~\ref{table:ikepublic}, KeyBot exhibits superior performance compared to the baseline models across all three datasets.
For instance, on the AASCE dataset, our proposed model nearly halves the NoC required to achieve a target MRE of 50 ($\text{NoC}_{10}$@50) compared to that of the baseline models.
This substantial reduction in NoC underscores KeyBot's proficiency in interactive frameworks, significantly lowering annotation costs.

\begin{figure}[t!]
\CenterFloatBoxes
\begin{floatrow}
\ffigbox
{\includegraphics[width=0.5\textwidth]{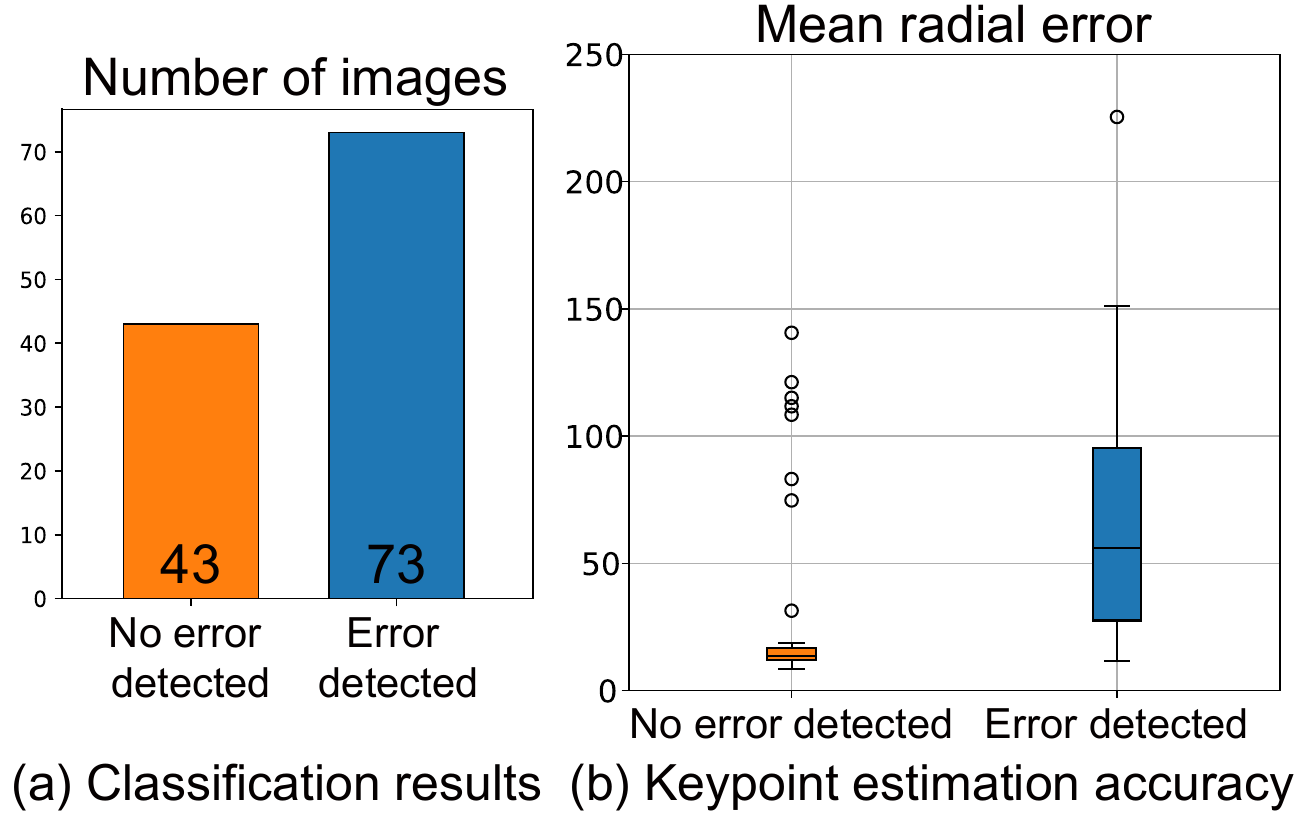}}
  {\caption{Effectiveness of the detector in identifying errors on the AASCE dataset.}\label{fig:suggestboxplot}}
 \killfloatstyle
\ttabbox
  {
      \centering
      \resizebox{0.35\textwidth}{!}{%
        \begin{tabular}{l|cc}
            \toprule
            \multirow{1}{*}{Method}
            & \makecell{No error \\ detected}  & \makecell{Error\\ detected}  \\
            \midrule
            \midrule
            Initial prediction     & 28.99 & 64.88 \\
            \midrule
            \ReVert\iterone     & - & 53.12  \\
            \ReVert\itertwo     & - & 50.49 \\
            \ReVert\iterthree     & - & 49.18 \\
            \bottomrule
        \end{tabular}
        }
  }
  {\caption{Keypoint prediction errors on the AASCE dataset, comparing MRE for examples where the detector identifies errors versus where it does not.}\label{tab:suggesterror}}
\end{floatrow}
\end{figure}

\noindent \textbf{Analysis on the detector.}
We investigate the effectiveness of the detector in identifying erroneous keypoints within images. 
In Fig.~\ref{fig:suggestboxplot}, we provide (a) the number of images where the detector identifies no errors versus those where errors are detected, as well as (b) the distribution of their actual mean radial error. 
Out of the total images analyzed, 43 images have no errors detected, while 73 images have errors identified. The images classified as ``No error detected'' exhibit significantly lower errors compared to those classified as ``Error detected.''
These findings indicate the precision and effectiveness of the detector in distinguishing between erroneous predictions and accurate ones.  

We further examine error reduction through successive KeyBot iterations for images where errors are initially detected. 
As demonstrated in Table~\ref{tab:suggesterror}, KeyBot effectively enhances the accuracy of these error-detected images, significantly reducing errors in subsequent iterations. Notably, after three iterations of KeyBot, the MRE significantly decreases from 64.88 to 49.18. 
This reduction highlights KeyBot's ability to iteratively identify inaccuracies and improve predictions.

\noindent \textbf{Ablation study.}
This section validates each component of KeyBot.
First, we study the role of the detector under two scenarios: (i) employing corrector outputs directly as the final results and (ii) utilizing all corrector outputs as pseudo-corrections for the interaction model.
As detailed in Table~\ref{table:ablation},
both scenarios exhibit significant performance degradation compared to KeyBot, underscoring the importance of using the detector to selectively apply corrector outputs. 

Additionally, utilizing only the interactive keypoint estimation network, without the support of KeyBot, results in a significant performance degradation compared to the refined predictions with KeyBot.
When ablating the error types simulated during the training of the corrector, our findings show that training the corrector to recognize and rectify three distinct types of errors significantly improves its prediction accuracy compared to when it is trained on only one error type. This finding highlights the necessity of incorporating all three typical errors into the training process.

\begin{table}[t!]
\caption{{Ablation study of our keypoint refinement method on the \spinewebdata dataset.}}
\label{table:ablation}
\begin{center}
\resizebox{0.8\textwidth}{!}{%
\begin{tabular}{cccc||ccccc}
\toprule
\multicolumn{1}{c}{\multirow{2.3}{*}
{\makecell{Interaction \\ model}}} &  
\multicolumn{3}{c||}{\multirow{1}{*}
{\makecell{\KeyBot}}}&
\multicolumn{5}{c}{\multirow{1}{*}
{\makecell{Mean radial error}}}
\\
\cmidrule(l{2pt}r{2pt}){2-4} \cmidrule(l{2pt}r{2pt}){5-9}
&
\multicolumn{1}{c}{\multirow{1}{*}{\makecell{\Discriminator}}}&
\multicolumn{1}{c}{\multirow{1}{*}{\makecell{\Corrector}}}&
\multicolumn{1}{c||}{\multirow{1}{*}{\makecell{Error type}}}
&     \uczero &     \ucone &    \uctwo &     \ucthree &     \ucfour \\
\midrule
\midrule
\rowcolor{lightblue}
\cmark&\cmark&\cmark&all
    &  \textbf{44.18} &  \textbf{25.93} &  \textbf{20.66} &  \textbf{18.03} &  \textbf{16.37} \\
\midrule
\xmark&\xmark&\cmark&all
    &  50.99 &  39.17 &  32.98 &  30.36 &  28.57 \\
\midrule
\cmark&\xmark&\cmark&all
    &  57.61 &  45.94 &  39.11 &  36.01 &  33.26 \\
\midrule
\cmark&\xmark&\xmark&-
    &  51.58 &  30.60 &  25.57 &  21.07 &  18.55 \\
\midrule
\cmark&\cmark&\cmark&misvertex
    &  52.67 &  31.14 &  25.92 &  21.65 &  18.52 \\
\cmark&\cmark&\cmark&misbone
    &  49.07 &  29.85 &  23.43 &  20.75 &  18.05 \\
\cmark&\cmark&\cmark&lr-inversion
    &  52.38 &  30.16 &  25.13 &  20.90 &  18.14 \\
\bottomrule
\end{tabular}
}
\end{center}
\end{table}

\begin{figure}[t!]
\includegraphics[width=1.0\textwidth]{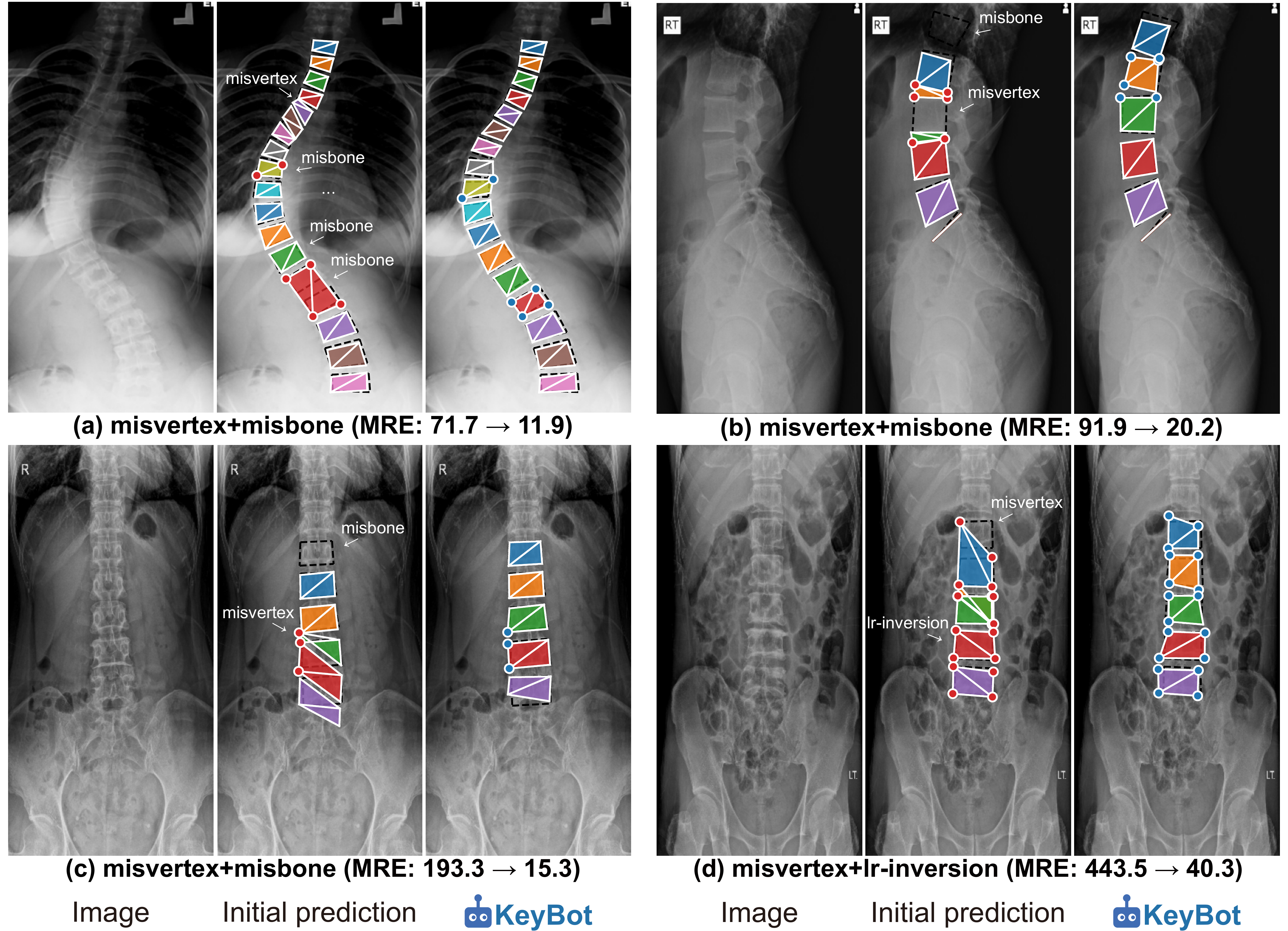}
\caption{Prediction results on the (a) 
 \spinewebdata, (b) \buudataLA, and (c,d) \buudataAP (bottom) datasets. KeyBot effectively reduces all three types of errors, even in challenging cases where error types are mixed. 
 The keypoints classified as inaccurate by our detector are indicated with red dots, while their corresponding keypoints in the revised results are visualized as blue dots. A diagonal line connecting the upper-right vertex to the lower-left vertex is indicated in each vertebra. More results are in Appendix~\ref{appendix:morequal}.
 } \label{fig:qual}
\end{figure}

\noindent \textbf{Qualitative results.}
In Fig.~\ref{fig:qual}, we analyze the qualitative results of KeyBot with three iterations. 
The results demonstrate that KeyBot effectively identifies and reduces all three error types: \misvertex (Misvertex), \misbone (Misbone), and \lrinv (LR-inversion). For these errors, KeyBot provides corrective feedback to the interactive model, enabling necessary adjustments.
For instance, in Fig.~\ref{fig:qual}(a), the initial prediction skips a bone, resulting in a Misbone error (red), and also exhibits a Misvertex error (purple). KeyBot accurately refines these errors, significantly reducing the MRE from 71.7 to 11.9. 
The results highlight KeyBot's capability to significantly enhance keypoint estimation accuracy without user clicks.

Additionally, KeyBot excels in interpretability. It includes a separate detector that enables users to easily comprehend the keypoints identified as errors and the corrections made, leading to efficient evaluation of the results. In the given examples, identified errors are indicated by red dots, and the corrections are shown as blue dots, illustrating how the keypoints are adjusted.
This transparency contributes to improving the efficiency of the overall annotation process.

\begin{figure}[t!]
\centering
\includegraphics[width=1.0\textwidth]{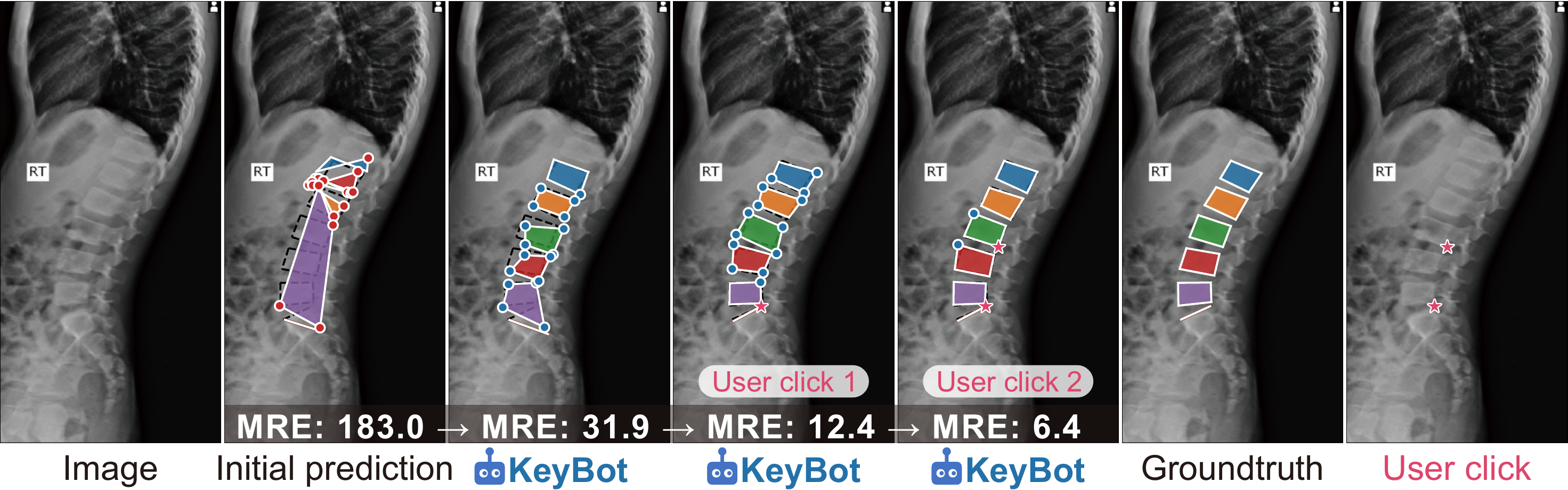}
\caption{Collaborative keypoint refinement process among the interaction model, KeyBot, and the user. Faced with substantial errors in initial predictions, KeyBot effectively removes all major errors, allowing users to easily identify and correct remaining inaccuracies with minimal effort. Here, refinement is achieved with just two user clicks.} \label{fig:qual_userhint}
\end{figure}

\noindent \textbf{Collaborative refinement process.}
In Fig.~\ref{fig:qual_userhint}, we showcase the qualitative results of the collaborative refinement process involving both KeyBot and the user.
Our approach establishes a feedback loop among the interaction model, KeyBot, and the user, combining expert medical knowledge and a targeted focus on specific error types. This significantly enhances the overall accuracy and efficiency of the annotation process.

\section{Conclusion} 
We introduce a novel approach, KeyBot, specifically designed to identify and correct significant errors in vertebrae keypoint estimation. 
Our proposed error simulation training procedure enables KeyBot to accurately and independently address specific error types, thereby significantly reducing the need for user intervention in the annotation process.
Comprehensive evaluations on public real-world datasets confirm the superiority of KeyBot over existing methods.

\section*{Acknowledgments}
This work was supported by Institute for Information \& communications Technology Promotion (IITP) grant funded by the Korea government (MSIT) (No.RS-2019-II190075 Artificial Intelligence Graduate School Program (KAIST)) and the National Research Foundation of Korea (NRF) grant funded by the Korea government (MSIT) (No. NRF-2022R1A2B5B02001913).

\bibliographystyle{splncs04}
\bibliography{main}

\renewcommand{\thesection}{\Alph{section}}
\renewcommand{\thesubsection}{\thesection.\arabic{subsection}}

\title{Supplementary Materials for \texorpdfstring{\\}{} Bones Can't Be Triangles: \texorpdfstring{\\}{} Accurate and Efficient Vertebrae Keypoint Estimation through Collaborative Error Revision}
\author{}
\authorrunning{J.~Kim et al.}
\titlerunning{Bones Can't Be Triangles}
\institute{}

\maketitle

\setcounter{table}{4}
\renewcommand{\thetable}{\arabic{table}}
\renewcommand{\thefigure}{\arabic{figure}}
\setcounter{figure}{8}

\begin{figure}[b!]
\includegraphics[width=\textwidth]{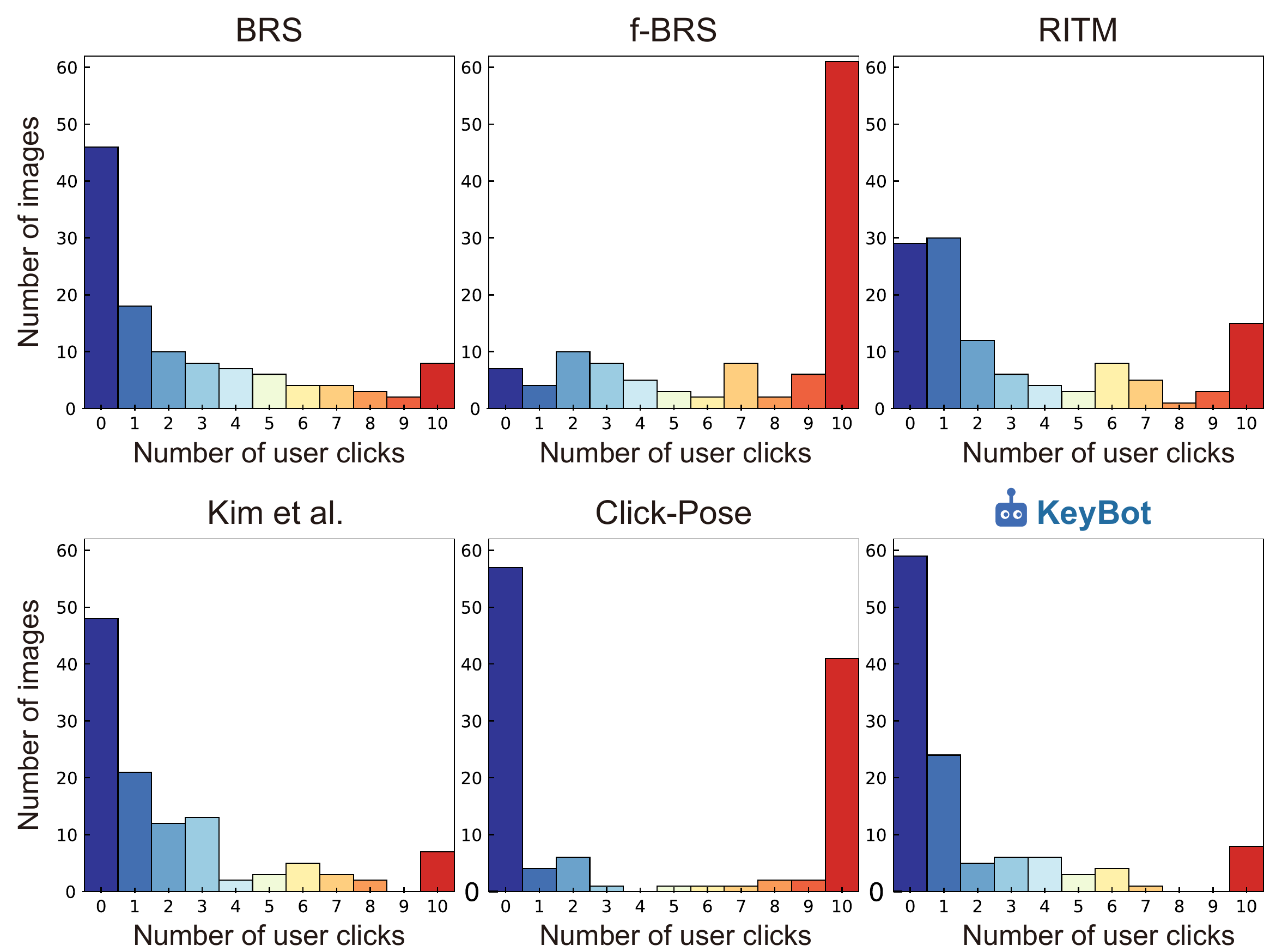}
\caption{Distribution of the number of user clicks (NoC) necessary to reach a target mean radial error (MRE) of 20 on the \spinewebdata dataset.}\label{supple_fig:nocdist}
\end{figure}

\noindent
\textbf{Summary.} This supplementary material enriches the main manuscript by providing comprehensive details of our methodology, additional visualizations, and extended experimental results. 
Section~\ref{appendix:morequan} presents additional experimental analyses, including a study of user click distribution, the integration of KeyBot with various interaction models, training KeyBot with real keypoint errors, annotation time comparison, and sensitivity analysis. 
Section~\ref{appendix:moremodeldetails} offers an in-depth description of our approach, outlining both its conceptual framework and algorithm.
Section~\ref{appendix:moreexpdetails} elaborates on the implementation details, experimental setups, and baselines used for comparison. 
Section~\ref{appendix:beam} investigates the application of KeyBot in a context where multiple refinement paths are explored, and the best path is chosen by the user, demonstrating its enhanced utility and effectiveness in the keypoint annotation process.
Section~\ref{appen:discussion} discusses the limitations and broad impact of our work.
Section~\ref{appendix:morequal} provides additional qualitative results, further affirming the practicality and impact of our work.

\section{More experimental results}~\label{appendix:morequan}
Section~\ref{appen:sec:1} analyzes the distribution of the number of user clicks. 
Section~\ref{appen:sec:2} explores the integration of KeyBot with various interaction models.
Section~\ref{appen:realerror} investigates training KeyBot with real keypoint errors.
Section~\ref{appen:sec:3} conducts a sensitivity analysis on the detector.

\subsection{Comparative analysis of user clicks required for target accuracy}\label{appen:sec:1}
In this analysis, we examine the distribution of the number of user clicks (NoC) required to achieve a target mean radial error (MRE) of 20, denoted as $\text{NoC}_{10}$@20. 
We assess the performance of our proposed method, KeyBot, which operates with a maximum of three iterations, in comparison to various baseline models, as shown in Fig.~\ref{supple_fig:nocdist}.
Notably, KeyBot reaches the target MRE with zero user clicks for a significant proportion of images, outperforming the baseline models.
In more than half of the entire instances, the images achieve the required accuracy autonomously, demonstrating KeyBot's efficiency in preemptively correcting major errors. 
KeyBot predominantly achieves the target MRE with fewer user interactions, reinforcing its effectiveness in reducing user effort while maintaining high accuracy in vertebrae keypoint estimation.

\subsection{Enhancing interactive keypoint estimation with keyBot}\label{appen:sec:2}
We rigorously assess KeyBot by integrating it with two interactive keypoint estimation frameworks, the model proposed by Kim et al.~\cite{kim2022morphology_mike} and Click-Pose~\cite{yang2023neural_clickpose}. 
The results, shown in Table~\ref{supple_tab:modelagnostic}, include:

\noindent\textbf{Manual revision (\texttt{manual}).} Manual correction refers to user adjustments on the initial predictions of Kim et al. and Click-Pose without any subsequent model modification, assessing error reduction achieved solely through user revision.

\noindent\textbf{Model revision.} The model revision results in gray show improvements due to automated refinements based on user feedback. 
The difference from manual revisions quantifies the error reduction attributable to the interaction model.

\noindent\textbf{KeyBot.} The KeyBot results in blue demonstrate its performance when added to existing models. The same KeyBot instance integrates seamlessly with either framework, requiring no additional training. 

\noindent\textbf{KeyBot without accumulating false predictions (\texttt{w/o fp}).} The results show the performance of KeyBot without the proposed false prediction accumulation strategy, specifically with the Kim et al. model, because this aspect is not applicable to Click-Pose (refer to Section~\ref{appen_sec_reproduce} for more details).

Our findings reveal that adopting KeyBot consistently improves the keypoint estimation performance of baseline models across all three datasets, demonstrating its robustness and effectiveness.
KeyBot significantly reduces errors in scenarios with no user interaction. 
For instance, on the AASCE dataset, KeyBot notably lowers MREs, a trend also observed in the BUU-AP and BUU-LA datasets. 
The model demonstrates further error reduction post user interactions. Its variant (\texttt{w/o fp}) exhibits slightly lower, yet still notable, performance.
Overall, the results demonstrate that KeyBot is model-agnostic and can be adapted to different frameworks without the need for model-specific training.

\begin{table*}[t!]
\caption{Performance comparison of mean radial error in keypoint estimation across the AASCE, BUU-AP, and
BUU-LA datasets. \texttt{UC} denotes the count of user clicks provided to the model.}
\label{supple_tab:modelagnostic}
\begin{center}
\resizebox{1.0\textwidth}{!}{%
\begin{tabular}{l|c|ccccc|ccc|ccc}
\toprule
\multicolumn{1}{l}{\multirow{2.3}{*}{Method}} & 
\multicolumn{1}{|c|}{\multirow{2.3}{*}{\makecell{Interaction\\backbone}}} & 
\multicolumn{5}{c}{\spinewebdata} &
\multicolumn{3}{c}{\buudataAP} &
\multicolumn{3}{c}{\buudataLA} \\
\cmidrule(l{2pt}r{2pt}){3-7} \cmidrule(l{2pt}r{2pt}){8-10}\cmidrule(l{2pt}r{2pt}){11-13}
 &  &
\makecell{\uczero} &\makecell{\ucone} &\makecell{\uctwo} &\makecell{\ucthree} &\makecell{\ucfour} &
\makecell{\uczero} &\makecell{\ucone} &\makecell{\uctwo} &
\makecell{\uczero} &\makecell{\ucone} &\makecell{\uctwo} \\
\midrule\midrule
\baselineikename~\cite{kim2022morphology_mike} &\texttt{manual}
     & 51.58 & 49.73 & 48.07 & 46.49 & 44.98 
    & 42.31 & 38.65 & 35.48
    & 23.43 & 20.81 & 18.63 
\\
\rowcolor{Gray}\baselineikename~\cite{kim2022morphology_mike} 
&\baselineikename
    & 51.58 & 30.60 & 25.78 & 21.60 & 19.08
    & 42.31 & 23.26 & 16.46
    & 23.43 & 14.29 & 10.29
\\
KeyBot\iterthree \texttt{w/o fp} &\baselineikename
    & 43.54 & 27.15 & 23.36 & 19.34 & 16.63 
    & \textbf{31.85} & \textbf{20.65 }& 15.87
    & 18.97 & \textbf{13.34} & 9.00
\\
\rowcolor{lightblue}{KeyBot\iterone}
&\baselineikename
    & 44.18 & \textbf{25.93} & \textbf{20.66} & \textbf{18.03} & 16.37
    & 32.01 & 21.79 & 15.97
    & 18.77 & 13.39 &  9.12
\\
\rowcolor{lightblue}{KeyBot\itertwo}
&\baselineikename
    & 42.52 & 27.03 & 22.95 & 18.72 & \textbf{16.23}
    & 31.88 & \textbf{20.65} & \textbf{15.81}
    & 19.11 & 13.47 &  9.03
\\
\rowcolor{lightblue}{KeyBot\iterthree}
&\baselineikename
    & \textbf{41.70} & 26.59 & 25.02 & 21.23 & 16.76
    & {31.87} & 20.66 & 15.84
    & \textbf{18.74} & {13.36} &  \textbf{8.97}
\\
\midrule  
Click-Pose~\cite{yang2023neural_clickpose} 
&\texttt{manual}
     & 54.65 & 52.80 & 51.27 & 49.85 & 48.48
     & 32.72 & 29.25 & 26.13 
     & 33.70 & 30.02 & 26.88
\\
\rowcolor{Gray}{Click-Pose~\cite{yang2023neural_clickpose} } 
&Click-Pose
    & 54.65 & 46.50 & 44.08 & 41.73 & 40.04
    & 32.72 & 29.30 & 26.20    
    & 33.70 & 21.62 & 17.38
\\
\rowcolor{lightblue}{KeyBot\iterone}
&Click-Pose
    & 52.62 & 46.10 & 43.65 & 41.17 & 39.39
    & 31.66 & 28.03 & 24.50
    & 33.98 & 20.53 & \textbf{16.87}
\\
\rowcolor{lightblue}{KeyBot\itertwo}
&Click-Pose
    & 51.38 & 45.98 & 43.53 & 40.98 & 39.09
    & \textbf{31.45} & \textbf{27.91} & \textbf{24.27}
    & 33.29 & 20.31 & 16.97 
\\
\rowcolor{lightblue}{KeyBot\iterthree}
&Click-Pose
    & \textbf{51.24} & \textbf{45.92} & \textbf{43.48} & \textbf{40.89} & \textbf{38.87}
    & {31.56} & {28.01} & {24.35}
    & \textbf{32.39} & \textbf{20.19} & {17.14 }
\\
\bottomrule
\end{tabular}
}
\end{center}
\end{table*}

\subsection{Training KeyBot with real keypoint errors}\label{appen:realerror}
We experiment with including real keypoint mistakes in the KeyBot training dataset, using a probability distribution of 40\% real mistakes, 40\% synthetic errors, and 20\% accurate keypoints. 
However, including real errors resulted in decreased performance, as shown in Table~\ref{table:realerror}. 
Real errors have high variability and lack consistent patterns, making it challenging to identify a clear pattern to correct. 
In contrast, synthetic errors are clearly defined and consistent, facilitating better convergence during training and leading to improved performance.
\begin{table}[!t]
\caption{Comparison of real and synthetic errors on the AASCE dataset.}
\label{table:realerror}
\begin{center}\resizebox{0.75\linewidth}{!}{
\begin{tabular}{cc|ccccc|cccc}
\toprule
\multicolumn{2}{c|}{Training} &\multicolumn{5}{c|}{Mean radial error} 
&\multirow{2.4}{*}{\makecell{$\text{NoC}_{10}$\\@20}} & \multirow{2.4}{*}{\makecell{$\text{NoC}_{10}$\\@30}} & \multirow{2.4}{*}{\makecell{$\text{NoC}_{10}$\\@40}} & \multirow{2.4}{*}{\makecell{$\text{NoC}_{10}$\\@50}} \\
\cmidrule(l{2pt}r{2pt}){1-2} \cmidrule(l{2pt}r{2pt}){3-7} 
syn & real & - & \texttt{UC1} & \texttt{UC2} & \texttt{UC3} & \texttt{UC4} \\
\midrule
 \xmark & \xmark  & 51.58 & 30.60 & 25.78 & 21.60 & 19.08 &  2.10 &  1.54 &  1.23 &  1.03 \\
\cmark & \xmark  & \textbf{41.70} & \textbf{26.59} & \textbf{25.02} & 21.23 & \textbf{16.76} &  \textbf{1.74} &  \textbf{1.32} &  \textbf{0.94} &  \textbf{0.68}  \\
\cmark &  \cmark  & 46.42 & 28.64 & 26.09 & \textbf{19.89} & 17.80 &  1.88 &  1.34 &  0.99 &  0.72  \\
 \xmark &  \cmark & 51.55 & 30.64 & 26.44 & 21.14 & 18.37 &  2.11 &  1.52 &  1.28 &  1.01  \\
\bottomrule
\end{tabular}}
\end{center}
\end{table}

\subsection{Comparison of annotation time}\label{appen:userstudy}
We conduct a user study with 15 participants, each tasked with annotating 22 keypoints on ten challenging radiographs from the BUU-LA dataset~\cite{klinwichit2023buu}. Participants are divided into three groups: one using KeyBot, one without it, and one using only initial model predictions without subsequent model revision.
As summarized in Table~\ref{table:annotation}, the results show that KeyBot significantly reduces annotation time and user clicks, demonstrating its efficiency.
Although the computation time for KeyBot is higher, its average inference time remains under $0.22$ seconds, which is negligible.

\begin{table}[!t]
\footnotesize
\caption{Comparison of annotation cost per image on the BUU-LA dataset.}
\label{table:annotation}
\begin{center}
\begin{tabularx}{0.68\textwidth}{l|YY}
\toprule
Method & Time (s) & {User click} \\
\midrule
manual revision  &  23.10 \scriptsize{$\pm$ 2.93} & 7.02 \scriptsize{$\pm$ 1.74}  \\
model revision w/o KeyBot \ \  & \  21.61 \scriptsize{$\pm$ 12.66} & 3.04 \scriptsize{$\pm$ 1.94}  \\
model revision w/ KeyBot  &  \textbf{11.71} \scriptsize{$\pm$ 2.23} & \textbf{0.44} \scriptsize{$\pm$ 0.30}  \\
\bottomrule
\end{tabularx}
\end{center}
\end{table}

\begin{table}[t!]
\caption{{Sensitivity analysis of the detector. MRE is measured on the BUU-AP and
BUU-LA datasets. $k$ and $s$ denotes represents the number of keypoints examined simultaneously and the stride during inference, respectively. \texttt{UC} denotes the number of user clicks. KeyBot\iterthree is used for the analysis.
}}
\label{sup_table:detectorsen}
\begin{center}
\begin{tabularx}{0.6\textwidth}{YY||ccc|ccc}
\toprule
\multicolumn{2}{c||}{Detector}
&\multicolumn{3}{c}{{\buudataAP}}
&\multicolumn{3}{c}{{\buudataLA}}\\
\cmidrule(l{2pt}r{2pt}){1-2}
\cmidrule(l{2pt}r{2pt}){3-5}\cmidrule(l{2pt}r{2pt}){6-8}
\multirow{1}{*}{\makecell{$k$}} 
&\multirow{1}{*}{\makecell{$s$}}
&\makecell{\uczero} &\makecell{\ucone} &\makecell{\uctwo}
&\makecell{\uczero} &\makecell{\ucone} &\makecell{\uctwo}
\\
\midrule
\midrule
$68$&   -
    & 39.32 & 22.39 & 19.64 
    & 20.44 & 13.94 &  9.53
    \\
\midrule
8&1 
    &  31.87 &  18.30 &  16.02 
    & \textbf{18.61} & 14.87 & \textbf{8.94} 
    \\
8&2 
    &  34.85 &  18.70 &  15.97
    & 18.64 & 13.44 & 8.96
    \\
8&3 
    &  \textbf{31.75}& \textbf{18.25} &  15.94 
    & 18.74 & 13.44 & 8.99 
    \\
\multicolumn{1}{c}{{8}} &    4 
    & 31.87 & 20.66 & \textbf{15.84}
    & 18.74 & \textbf{13.36} & 8.97 
    \\
\midrule
4&1
    & 36.17 & 20.25 & 16.24
    & 18.79 & 13.42 &  8.99
\\
4&2
    & 32.22 & 20.86 & 16.34
    & \textbf{18.42} & \textbf{13.39} &  9.07
\\ 
4&3
    & \textbf{31.74} & \textbf{18.14} & \textbf{16.03}
    & 18.56 & 13.46 & 10.07
\\ 
$4$&   4
    & 32.27 & 22.94 & 16.14
    & {18.61} & 15.10 &  \textbf{8.95}
    \\
\bottomrule
\end{tabularx}
\end{center}
\end{table}

\subsection{Sensitivity analysis of the detector in keypoint error detection}\label{appen:sec:3}

We conduct a sensitivity analysis on the detector, as detailed in Table~\ref{sup_table:detectorsen}.
We investigate the impact of varying the number of simultaneously examined keypoints ($k$) and the stride ($s$) during inference on keypoint estimation accuracy.
We observe a substantial decrease in performance when the detector assesses all keypoints at once, resulting in the most significant keypoint estimation errors. 
This performance decline suggests that analyzing the entire bone structure at once increases complexity, diminishing the detector's effectiveness. Focusing on specific bone segments during detection proves to be more efficient. 
 KeyBot maintains robust performance with different input keypoint numbers, especially when $k=4$ and $k=8$.

Additionally, we analyze the effect of varying the stride $s$ with a fixed keypoint window $k$. KeyBot performs consistently across different stride settings. Particularly with an input keypoint number of four, smaller stride values than $k$ slightly enhance performance. A smaller stride provides a more detailed and contextually varied examination of each keypoint, improving the detector's accuracy and reliability.

\begin{figure}[t!]
\includegraphics[width=\textwidth]{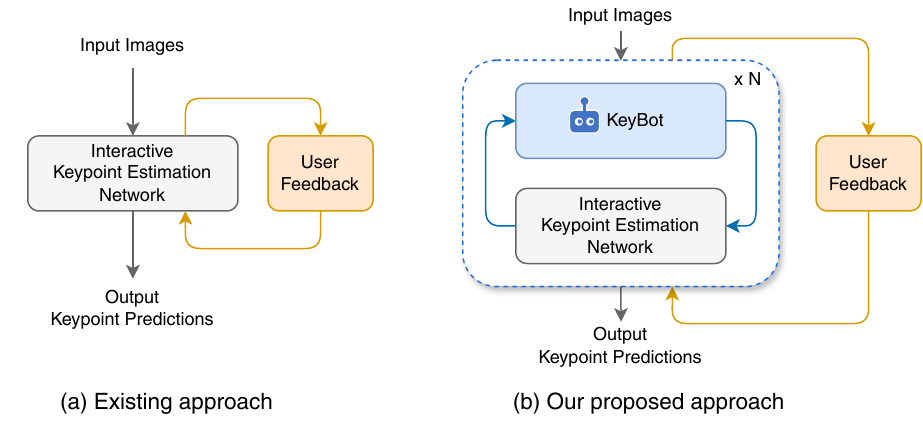}
\caption{Comparison between the (a) existing interactive keypoint estimation framework and (b) the proposed approach adopting KeyBot.} \label{supple_fig:flow}
\end{figure}

\section{Additional details of KeyBot}~\label{appendix:moremodeldetails}
This section complements an explanation of KeyBot. First, Section~\ref{appen:flow} describes the detailed conceptual framework of our method, and Section~\ref{appen:algorithm} offers an algorithm that elucidates the overall process of KeyBot.

\subsection{Overview of the proposed approach}~\label{appen:flow}
Existing interactive keypoint estimation approaches~\cite{kim2022morphology_mike, yang2023neural_clickpose} operate by predicting keypoints from images and refining these predictions based on user feedback, as depicted in Fig.~\ref{supple_fig:flow}(a). However, they lack the capability for self-correction without user intervention.

To address this limitation, we introduce KeyBot, which independently evaluates the interaction model's predictions, pinpointing and correcting errors autonomously, as illustrated in Fig.~\ref{supple_fig:flow}(b).
This approach facilitates autonomous correction of predictions, thereby enabling users to concentrate on subsequent refinements after KeyBot's preliminary error rectification.

\subsubsection{Why a user-interactive approach?}
In medical imaging, the accuracy of keypoint estimation models is paramount. 
Despite recent advancements, human adjustments are often needed to address inherent model biases and errors.
However, manual revision, while necessary to ensure reliability and accuracy, is often a tedious and time-consuming task. 
Thus, an interactive approach incorporating human feedback aims to streamline this process by refining inaccuracies with minimal user intervention.

\subsubsection{Why KeyBot?}
Existing interactive models require users to initiate error assessment, which can be laborious and error-prone, especially with numerous predictions or clustered inaccuracies. KeyBot autonomously conducts an initial assessment, addressing basic errors such as fundamental misidentifications of bone structures.
By addressing these primary errors, it reduces the time users spend on basic error corrections, allowing for a more focused and efficient use of human expertise.

\subsubsection{Why not end-to-end?}
While an interactive keypoint estimation model learns to correct its errors based on user feedback, it lacks explicit training for specific error types. In contrast, KeyBot is designed to identify and correct three specific error types, utilizing error simulation in its training phase. 
Its independent structure prevents it from inheriting potential biases or limitations of the interaction model, ensuring an objective and reliable keypoint estimation process.

\subsubsection{Ongoing collaborative loop} 
Integrating KeyBot within the interactive keypoint estimation framework in a feedback-providing manner maintains an iterative and collaborative loop. This loop ensures that the refinement process is not a one-off correction but a continuous improvement cycle, accommodating further refinements and adjustments from both KeyBot and human users, leading to more accurate outcomes in medical image analysis.

\subsection{Algorithm of KeyBot}\label{appen:algorithm}
The complete procedure of our proposed approach is encapsulated in Algorithm~\ref{algorithm}.
Given an input image, the interaction model makes an initial keypoint prediction, followed by two sequential phases: the KeyBot phase (with $N$ iterations) and the user phase (one iteration). 
During the KeyBot phase, KeyBot performs a preliminary step to correct errors and generates corrective feedback. This feedback is fed into the interaction model, akin to user feedback. The interaction model then generates a new, corrected prediction.  This phase repeats for $N$ iterations or until KeyBot detects no errors. Subsequently, in the user phase, the user corrects a single error, and the interaction model updates its results accordingly. This entire process repeats until it reaches the maximum number of user interactions, $T$.

\definecolor{lightblue}{rgb}{0.87, 0.92, 0.97}
\definecolor{lightpink}{rgb}{0.9803921568627451, 0.8784313725490196, 0.8470588235294118} 

\algrenewcommand{\algorithmiccomment}[1]{\hspace{1em}$\triangleright$ #1}

\begin{algorithm}[H]
\DontPrintSemicolon
\LinesNumberedHidden
\KwData{input image $\boldsymbol{x}$, maximum user click $T$, maximum \KeyBot iteration number $N$, interaction model $\mathcal{F}_\theta$, \KeyBot~\discriminator $\mathcal{G}_{\phi_r}$, \KeyBot~\corrector $\mathcal{G}_{\phi_s}$}
\KwResult{$\boldsymbol{y}$, the final keypoint prediction}
${\boldsymbol{c}_{0,0}=0, \boldsymbol{e}_{0,0}=0}$\;
\hspace*{-\fboxsep}\colorbox{Gray}{\parbox{0.95\linewidth}{%
$\boldsymbol{y}_{0,0} \gets \mathcal{F}_\theta(\boldsymbol{x},{\boldsymbol{c}_{0,0}}, {\boldsymbol{e}_{0,0}})$\Comment{\fontfamily{qcr}\selectfont\footnotesize{initial interaction model forward}}\;%
}}
\While{$t<T$}{
    \While{$n<N$}{
            \hspace*{-\fboxsep}\colorbox{lightblue}{\parbox{0.855\linewidth}{%
               ${\boldsymbol{\nu}_{t,n\Plus1}} \gets \mathcal{G}_{\phi_r}(\boldsymbol{x}, {\boldsymbol{y}_{t,n}})$\Comment{\fontfamily{qcr}\selectfont\footnotesize{KeyBot \discriminator}}\;
            }}
            \eIf{$|\boldsymbol{\nu}_{t,n\Plus1} \setminus \boldsymbol{\rho}_{t}| = 0$}{
                  \textbf{break}
                }{
            \hspace*{-\fboxsep}\colorbox{lightblue}{\parbox{0.81\linewidth}{%
                ${\boldsymbol{z}_{t,n\Plus1}} \gets \mathcal{G}_{\phi_s}(\boldsymbol{x}, {\boldsymbol{y}_{t,n}})$\Comment{\fontfamily{qcr}\selectfont\footnotesize{KeyBot \corrector}}\;
                $\boldsymbol{c}_{t,n\Plus1}^i \gets {\boldsymbol{z}_{t,n\Plus1}^i, i \in {\boldsymbol{\nu}_{t,n\Plus1}} \setminus \boldsymbol{\rho}_{t}}$\;
                $\boldsymbol{e}_{t,n\Plus1}^i \gets {\boldsymbol{y}_{t,n}^i, i \in {\boldsymbol{\nu}_{t,n\Plus1}} \setminus {\boldsymbol{\nu}_{t,:n}}\setminus \boldsymbol{\rho}_{t}}$\;
                ${\boldsymbol{y}_{t,n\Plus1}} \gets \mathcal{F}_\theta(\boldsymbol{x}, {\boldsymbol{c}_{t,n\Plus1}}, {\boldsymbol{e}_{t,n\Plus1}})$\Comment{\fontfamily{qcr}\selectfont\footnotesize{interaction model forward}}\;
            }}
            }
        }
            $\boldsymbol{c}_{t\Plus1,0}\gets0, \boldsymbol{e}_{t\Plus1,0}\gets\boldsymbol{e}_{t,n\Plus1}$\;
            \hspace*{-\fboxsep}\colorbox{lightpink}{\parbox{0.9\linewidth}{%
            $\boldsymbol{\rho}_{t\Plus1}$, $\boldsymbol{u}_{t\Plus1} \gets \Psi(\boldsymbol{x},\boldsymbol{y}_{t,n\Plus1},\boldsymbol{u}_{t})$\Comment{\fontfamily{qcr}\selectfont\footnotesize\footnotesize{User revision}}\;
             $\boldsymbol{c}_{t\Plus1,0}^i \gets {\boldsymbol{u}_{t\Plus1}^i, i \in \boldsymbol{\rho}_{t\Plus1}}$\;
            $\boldsymbol{e}_{t\Plus1,0}^i \gets {\boldsymbol{y}_{t,n\Plus1}^i, i \in \boldsymbol{\rho}_{t\Plus1}  }$\;
            ${\boldsymbol{y}_{t\Plus1,0}} \gets \mathcal{F}_\theta(\boldsymbol{x}, {\boldsymbol{c}_{t\Plus1,0}}, {\boldsymbol{e}_{t\Plus1,0}})$\Comment{\fontfamily{qcr}\selectfont\footnotesize{interaction model forward}}\;
        }}
}\label{algorithm}
\caption{Inference with \KeyBot}
\end{algorithm}

\section{Experimental details}~\label{appendix:moreexpdetails}
This section provides comprehensive details about the implementation details (Section~\ref{appen_sec:modeldetail}), dataset descriptions (Section~\ref{appen_sec:dataset}), metric definitions (Section~\ref{appen_sec_mre}), and reproducibility details for baseline models (Section~\ref{appen_sec_reproduce}).

\subsection{Implementation details of KeyBot}\label{appen_sec:modeldetail}

We describe a detailed experimental setup of our approach: the detector, the corrector, and the interaction model. 
Also, we elaborate on the error simulation methods employed in training the detector and the corrector, including \misvertex (misvertex), \misbone (misbone), and \lrinv (lr-inversion).

\subsubsection{Detector} 
The detector analyzes eight ($k=8$) keypoints simultaneously, classifying each as accurate or inaccurate.
Input X-ray images, cropped and resized around these keypoints to $128\times128$ dimensions, are concatenated with Gaussian keypoint heamtaps. 
The detector evaluates the abnormality likelihood for each keypoint using a sigmoid function. Training labels are generated by marking synthetically displaced keypoints as one (indicative of errors) and the rest as zero. 
During inference, the detector iteratively processes keypoints with a stride of four ($s=4$). On the BUU-LA datasets, this process is applied to only 20 keypoints, excluding the final two for error detection. Any keypoint with an anomaly probability above 0.5 is flagged as erroneous.

The detector employs a modified ResNet-18~\cite{resnet} architecture, adapted for combined image and heatmap inputs. 
The training process incorporates simulated \misvertex errors, displacing up to three keypoints per image for AASCE and four keypoint for BUU-AP and BUU-LA datasets, respectively. 
Selected keypoints shift up to four indices away from their original position, with wrapping for out-of-range indices. 
The training utilizes Binary Cross-Entropy (BCE) loss over 300 epochs with early stopping (zero patience) and an AdamW optimizer with a learning rate of $0.001$.

\subsubsection{Corrector} The corrector processes the entire image, resized to $256\times128$, alongside Gaussian keypoint heatmaps of matching resolution. 
It generates reconstructed keypoint locations as heatmaps, using a sigmoid function in the final layer. 
The architecture is based on DeepLab-v3 with a ResNet50 encoder. 
During training, KeyBot is trained on accurate keypoints with a $20\%$ probability and simulated errors with an $80\%$ probability.
Training uses three error types or accurate keypoints, with varied probabilities for each dataset (Please refer to the source code for more details).
\begin{itemize}
\item[(1)]\textbf{Vertex misidentification errors}: Up to nine keypoints are displaced, with a multinomial probability distribution for the number of keypoints to shift. 
Keypoints are selected randomly with equal probability, and they are shifted to maximally four indices away from their original index.
    
\item[(2)]
\textbf{Bone misidentification errors}: To simulate misbone errors, movement type (up, down, or accurate) is selected with equal probability. For the shifts, there is an equal probability for each of the following scenarios: moving all keypoints from the first to the last, moving keypoints from a random starting point to the last keypoint, moving keypoints from the first to a random end point, and moving keypoints between a random start and end point.
    For keypoints located at either the first or last vertebrae, relocation may occur outside the targeted bone structure, with the magnitude determined by positional differences between either the first and second vertebrae or the last and its immediate predecessor.
    
\item[(3)]
\textbf{Left-right inversion errors}: Every left-right pair is independently swapped with a 90\% probability. 
\end{itemize}

\noindent
Similar to the detector, the corrector is trained over 300 epochs with early stopping, using an AdamW optimizer with the learning rate of $0.001$. 

\begin{table*}[t!]
\caption{Summary of X-ray image datasets for keypoint estimation used in our work.}
\label{appen_table:datainfo}
\begin{center}
\resizebox{1.0\linewidth}{!}{
\begin{tabular}{lccccccccc}
\toprule
 \multicolumn{1}{l}{\multirow{2.4}{*}{Dataset}}
& \multicolumn{1}{c}{\multirow{2.4}{*}{\makecell{Target \\keypoints}}} 
& \multicolumn{4}{c}{Number of images}
& \multicolumn{4}{c}{Human annotation error}\\
\cmidrule(lr){3-6}\cmidrule(lr){7-10}
 \multicolumn{1}{c}{}& \multicolumn{1}{c}{}
& Total
& \makecell{Train} 
& \makecell{Val} 
& \makecell{Test}
& Total
& \makecell{misvertex}
& \makecell{misbone} 
& \makecell{lr-inversion}\\
\midrule
\midrule
\spinewebdata~\cite{spinewebdataset} &  68
& 563
& 325 & 122 & 116
& 45 
& 18
& 27 
& -
\\
\buudataAP~\cite{klinwichit2023buu} & 20   
& 399
& 240 & 80 & 79
& 1
& - 
& -   
& 1  
\\
\buudataLA~\cite{klinwichit2023buu} & 22 
& 397
& 237 & 80 & 80
& 3 
& - 
& - 
& 3 
\\
\bottomrule
\end{tabular}
}
\end{center}
\end{table*}

\begin{figure}[t!]
\includegraphics[width=\textwidth]{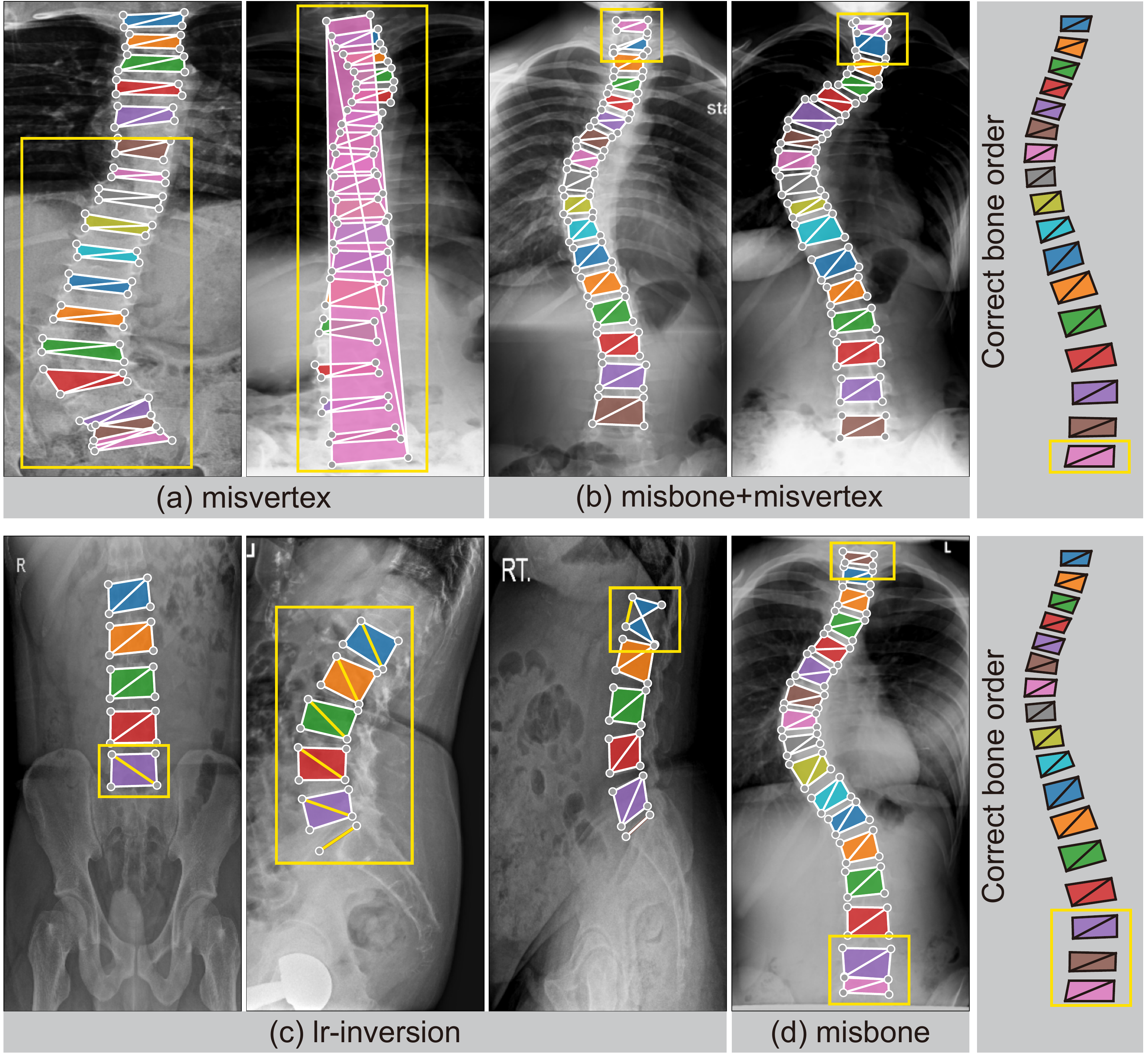}
\caption{Errors in human annotation present in the AASCE dataset in (a),(b) and (d), and in the BUU-AP and BUU-LA datasets in (c). A diagonal line connecting the upper-right vertex to the lower-left vertex is indicated in each vertebra.} \label{supple_fig:human}
\end{figure}

\subsubsection{Interaction model}
The interaction model estimates keypoint heatmaps from input X-ray images with a size of $512\times256$.
We extend the model proposed by Kim et al.~\cite{kim2022morphology_mike} by integrating revision feedback from either KeyBot or user input and adopting accumulated false predictions. 
Training includes iterative refinement based on simulated user feedback, providing groundtruth keypoint locations as user revision. During inference, the most significant keypoint error is modified to simulate user interaction.

\subsection{Datasets}\label{appen_sec:dataset}

In our study, we utilize public X-ray image datasets, namely AASCE, BUU-AP, and BUU-LA, as detailed in Table~\ref{appen_table:datainfo}.
An extensive analysis of these datasets reveals three primary types of human annotation errors: vertex misidentification (misvertex), bone misidentification (misbone), and left-right inversion (lr-inversion).
 The AASCE dataset predominantly exhibits misvertex and misbone errors, whereas the BUU-AP and BUU-LA datasets mainly exhibit lr-inversion errors. 
 To ensure the integrity and reliability of our evaluation, images with such critical annotation errors were excluded, as illustrated in Fig.~\ref{supple_fig:human}.

\subsubsection{\spinewebdata~\cite{spinewebdataset}}
The AASCE dataset consists of 608 anterior-posterior X-ray images, annotated with 68 keypoints across 17 vertebrae. 
We follow the original train-test split, further partitioning the training set for validation purposes. 
Error analysis identifies that it contains nine misvertex and 18 misbone errors in the training set, one misvertex and five misbone errors in the validation set, and eight misvertex and four misbone errors in the test set, as exemplified in Fig.~\ref{supple_fig:human}(a,b,d). Two cases showing a mixture of misvertex and misbone errors are counted as misbone errors.

\subsubsection{\buudataAP~\cite{klinwichit2023buu}}
Comprising 400 anterior-posterior view X-ray images, BUU-AP dataset is annotated with 20 keypoints per image. The image sizes range from $1434 \times 1072$ to $3072 \times 3040$. 
We randomly split the images into training, validation, and test sets, identifying a lr-inversion error in the test set, as shown in Fig.~\ref{supple_fig:human}(c).

\subsubsection{\buudataLA~\cite{klinwichit2023buu}}
The BUU-LA dataset includes 400 left lateral view X-ray images, each with 22 annotated spinal keypoints with image sizes ranging from $1956 \times 968$ to $3072 \times 3040$.
It undergoes similar partitioning as BUU-AP, with three lr-inversion errors identified in the training set, as exemplified in Fig.~\ref{supple_fig:human}(c).

\subsection{Definition of mean radial error}\label{appen_sec_mre}
In our study, we assess the keypoint estimation accuracy using the mean radial error (MRE).
MRE measures the average Euclidean distance between the predicted and groundtruth coordinates of keypoints.
Specifically, for each of the $K$ target keypoints in a sample, let $\boldsymbol{p}^*_i$ denote the 
groundtruth coordinates and $\boldsymbol{p}_i$ the predicted coordinates of the $i$-th keypoint.
The MRE is calculated as:
\begin{equation}
 \text{MRE}=\frac{1}{K}\sum_{i=1}^K\big|\big|\boldsymbol{p}^*_i-{\boldsymbol{p}}_i\big|\big|_2.
\end{equation}
MRE measures the overall precision of keypoint estimation results.

\subsection{Reproducibility for baselines}\label{appen_sec_reproduce}
 
\subsubsection{Interactive segmentation models.}
In this work, we compare our method with several interactive segmentation models, including BRS~\cite{brs}\footnote{https://github.com/wdjang/BRS-Interactive_segmentation}, f-BRS~\cite{fbrs}\footnote{https://github.com/saic-vul/fbrs_interactive_segmentation/tree/master}, and RITM~\cite{ritm}\footnote{https://github.com/SamsungLabs/ritm_interactive_segmentation}, for our keypoint estimation task.
Using their official source codes, we modify these models to produce outputs for the specific number of keypoints required in our study, aligning with the hyperparameter settings from Kim et al.~\cite{kim2022morphology_mike} for consistency.

\subsubsection{Kim et al.~\cite{kim2022morphology_mike}}
We adhere to the experimental settings outlined in the official source code\footnote{https://github.com/seharanul17/interactive_keypoint_estimation}. 
For experiments on the AASCE dataset, we utilize their pretrained network. 
In the case of the BUU-AP and BUU-LA datasets, we train the model using identical hyperparameters as those used for the AASCE dataset, with the only modification being the selection of keypoint subsets. This adjustment is necessary to accommodate the morphology-aware loss proposed in the study. 
When integrating KeyBot, we handle KeyBot's revision feedback as user interaction input, similar to processing user clicks.  

\subsubsection{Click-Pose~\cite{yang2023neural_clickpose}}
We utilize the official source code of Click-Pose\footnote{https://github.com/IDEA-Research/Click-Pose}, maintaining the hyperparameters consistent across all datasets, while adjusting the number of keypoints as required. 
During the training phase, we exclude the Object Keypoint Similarity (OKS) loss which is specifically tailored for human pose estimation.  
As a result, Click-Pose training involves using a combination of L1 loss for bounding boxes, intersection over union (IOU) loss, classification loss, and L1 loss for keypoints. Click-Pose employs a keypoint regression-based approach, directly estimating the coordinates of keypoints. For integration with KeyBot, we feed KeyBot's revision feedback into the human-to-keypoint decoder module of Click-Pose, akin to handling user clicks. Given that Click-Pose does not employ a heatmap-based keypoint estimation approach, we do not incorporate false predictions into the input framework. 

\begin{figure}[t!]
\includegraphics[width=\textwidth]{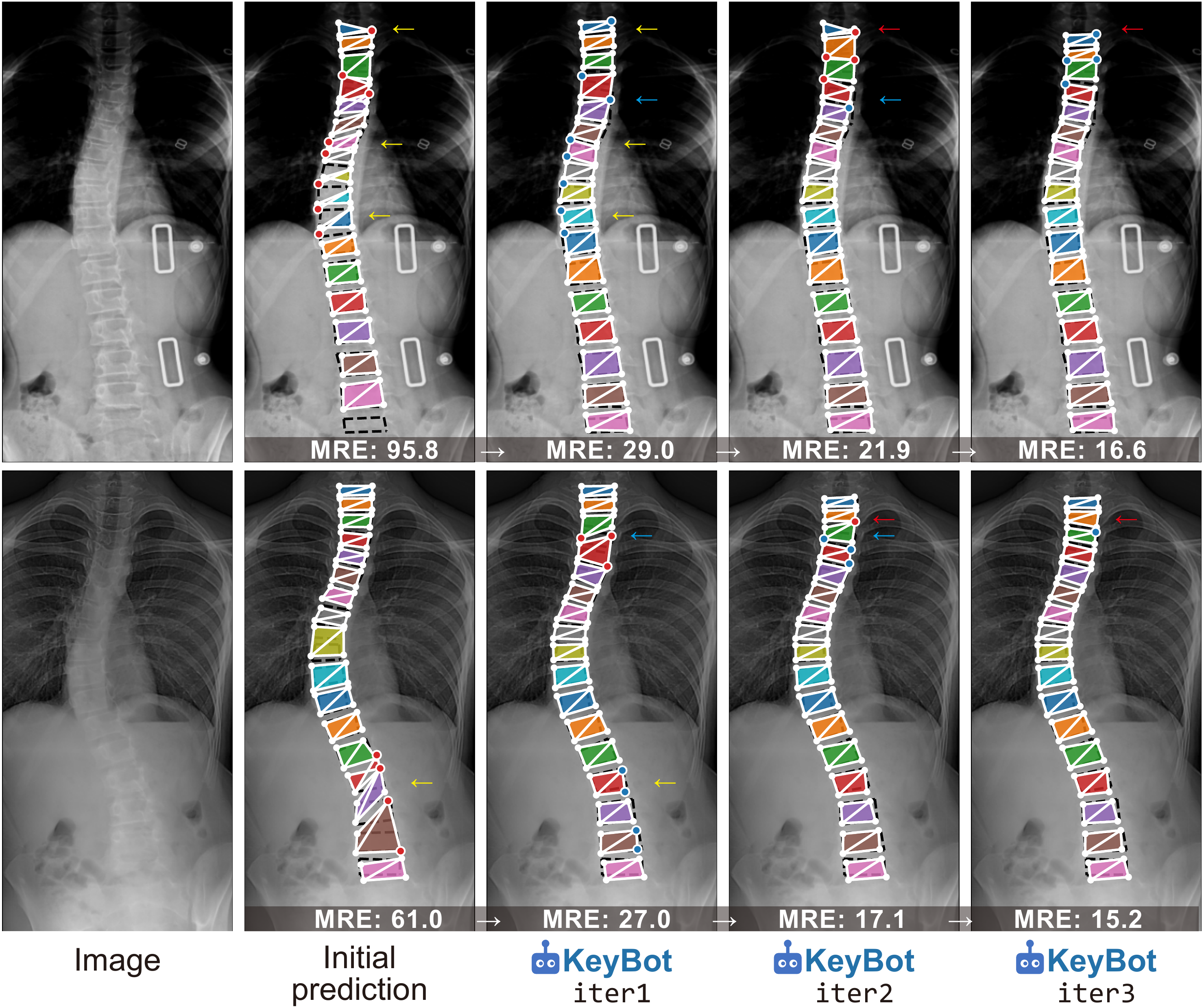}
\caption{Iterative refinement results of KeyBot on the AASCE dataset. In the first example, KeyBot corrects the upper and middle part in the image indicated by the yellow arrow, lowering the bones overall in the first iteration. 
It revises the blue arrow area in the second iteration and finally lowers the incorrectly positioned bones at the top in the third iteration. 
In the second example, the first iteration adjusts the lower part as indicated by the yellow arrow. In the second iteration, the front end of the bones marked in blue is lowered overall. Finally, in the third iteration, the model properly adjusts the position of the bones at the top, marked in red.
A diagonal line connecting the upper-right vertex to the lower-left vertex is indicated in each vertebra. 
} \label{supple_fig:beam}
\end{figure}

\section{Exploring multiple refinement paths of KeyBot}\label{appendix:beam}

\begin{table*}[t!]
\caption{Performance comparison of mean radial error in keypoint estimation across the AASCE, BUU-AP, and
BUU-LA datasets. For KeyBot\iterthree \texttt{w/ oracle}, the best prediction is chosen among three iterations. No user input is provided.}
\label{table:supple_beam}
\begin{center}
\resizebox{0.65\textwidth}{!}{%
\begin{tabular}{lc|c|c}
\toprule
\multicolumn{1}{l}{\multirow{1}{*}{Method}} & 
\multicolumn{1}{c|}{\spinewebdata} &
\multicolumn{1}{c|}{\buudataAP} &
\multicolumn{1}{c}{\buudataLA} \\
\midrule\midrule
\multicolumn{1}{l}{$\text{BRS}$~\cite{brs}} 
    & 45.65  
    & 51.22  
    & 40.20 
\\
\multicolumn{1}{l}{$\text{f-BRS}$~\cite{fbrs}}
    & 64.06 
    & 44.05  
    & 36.03 
\\
\multicolumn{1}{l}{RITM~\cite{ritm}} 
    & 56.03  
    & 36.43 
    & 23.27  
\\
\midrule  
\multicolumn{1}{l}{{\baselineikename~\cite{kim2022morphology_mike}}}
    & 51.58  
    & 42.31  
    & 23.43  
\\
\multicolumn{1}{l}{{Click-Pose~\cite{yang2023neural_clickpose}}} 
    & 54.65  
    & 32.72  
    & 33.70  
\\
\midrule  
\rowcolor{lightblue}\multicolumn{1}{l}{{KeyBot\iterone}}
    & 44.18  
    & 32.01 
    & 18.77 
\\
\rowcolor{lightblue}\multicolumn{1}{l}{{KeyBot\itertwo}}
    & 42.52  
    & 31.88  
    & 19.11  
\\
\rowcolor{lightblue}\multicolumn{1}{l}{{KeyBot\iterthree}}
    & {41.70}  
    & {31.87}  
    & {18.74} 
\\
\midrule
\rowcolor{lightblue}\multicolumn{1}{l}{{KeyBot\iterthree \texttt{w/ oracle}}}
    & \textbf{39.29}  
    & \textbf{31.43} 
    &  \textbf{18.70} 
\\
\bottomrule
\end{tabular}
}
\end{center}
\end{table*}

This section introduces a novel collaborative annotation approach involving KeyBot, the user, and the interaction model. We investigate the application of KeyBot in a context where multiple refinement paths are explored, and the best path is chosen by the user, demonstrating its enhanced utility and effectiveness in the keypoint annotation process. 
This method involves multiple iterative interactions between KeyBot and the backbone model, yielding a variety of refined results. These results, alongside the initial prediction, are presented to the user for selection and potential further refinement. Among multiple refinement iterations, users can select the most precise result as their foundation for any additional modifications. Users also have the option to ignore the model updates and correct them manually if needed. 

This approach resembles the concept of beam search, where each KeyBot refinement iteration represents a node in the search space. We allow users to explore multiple paths, i.e., KeyBot iterations, and to choose the best path.
Providing multiple iteration results is feasible because the average inference time for the interaction model is 0.181 seconds, and for KeyBot is 0.216 seconds, which is negligible.
The procedure consists of three steps. Initially, the backbone model produces an initial keypoint prediction from the input X-ray image, forming the baseline prediction.
Next, KeyBot evaluates this prediction, identifies inaccuracies, and makes corrections, leading to the first refined set of keypoints. 
This iteration repeats, resulting in multiple refined sets. Lastly, the user reviews all keypoint sets, including the initial and refined ones, and selects the most accurate set as a basis for any further adjustments.

The primary benefit of this approach is that it provides users with multiple refined results in addition to the initial prediction, as shown in Fig.~\ref{supple_fig:beam}. This allows for a well-informed decision-making process, where users select the most precise result as their foundation for any additional modifications.

Additionally, we analyze the error reduction achieved through this multi-iteration presentation approach, as shown in Table~\ref{table:supple_beam}. 
This multi-iteration presentation approach with KeyBot, combining KeyBot’s iterations with user selection and refinement, optimizes prediction accuracy while significantly reducing user effort.

\section{Discussion}\label{appen:discussion}
The anatomical complexity and similarity among vertebrae make vertebrae keypoint estimation prone to errors that require substantial human efforts to correct.
By advancing methodology to enhance accuracy and efficiency in this area, our work represents a substantial technical contribution.
However, a limitation is that its correction precision occasionally falls short of human-level precision when most keypoints contain substantial errors, as shown in Fig.~\ref{fig:qual_userhint} of our main manuscript.
In these cases, KeyBot disregards the erroneous predictions and generates completely new ones. However, because KeyBot generates revisions based on the initial predictions (refer to Fig.~\ref{fig:method} of our main manuscript), it is influenced by these errors, limiting its ability to provide precise feedback for incorrect vertebrae shapes. This necessitates user feedback to guide further precise adjustments. Future work aims to develop a more robust feedback mechanism to address these cases effectively, enhance accuracy, and minimize user input.

Although our approach is developed specifically for vertebrae, it introduces a general framework for addressing domain-specific errors that can be extended to other fields.
By characterizing domain-specific error types and generating synthetic data to represent these inaccuracies, our approach can be used to develop an auxiliary model to detect and correct them efficiently.
This adaptability highlights the broader impact and versatility of our work.

\section{Additional qualitative results}~\label{appendix:morequal}
\begin{figure}[t!]
\includegraphics[width=\textwidth]{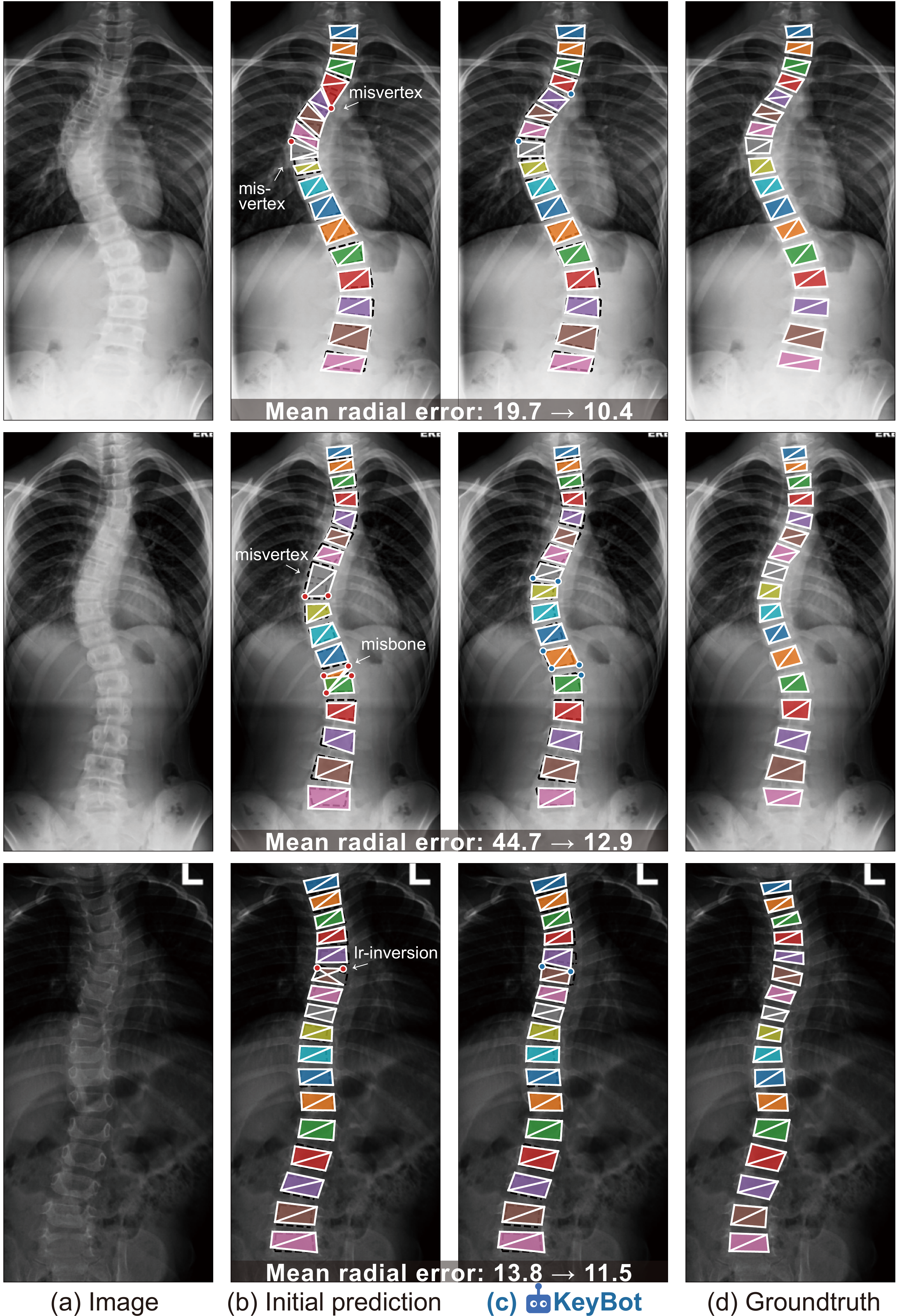}
\caption{Additional qualitative results of KeyBot on the \spinewebdata dataset, with a maximum of three iterations. 
} \label{supple_fig:qual1}
\end{figure}

\begin{figure}[t!]
\includegraphics[width=\textwidth]{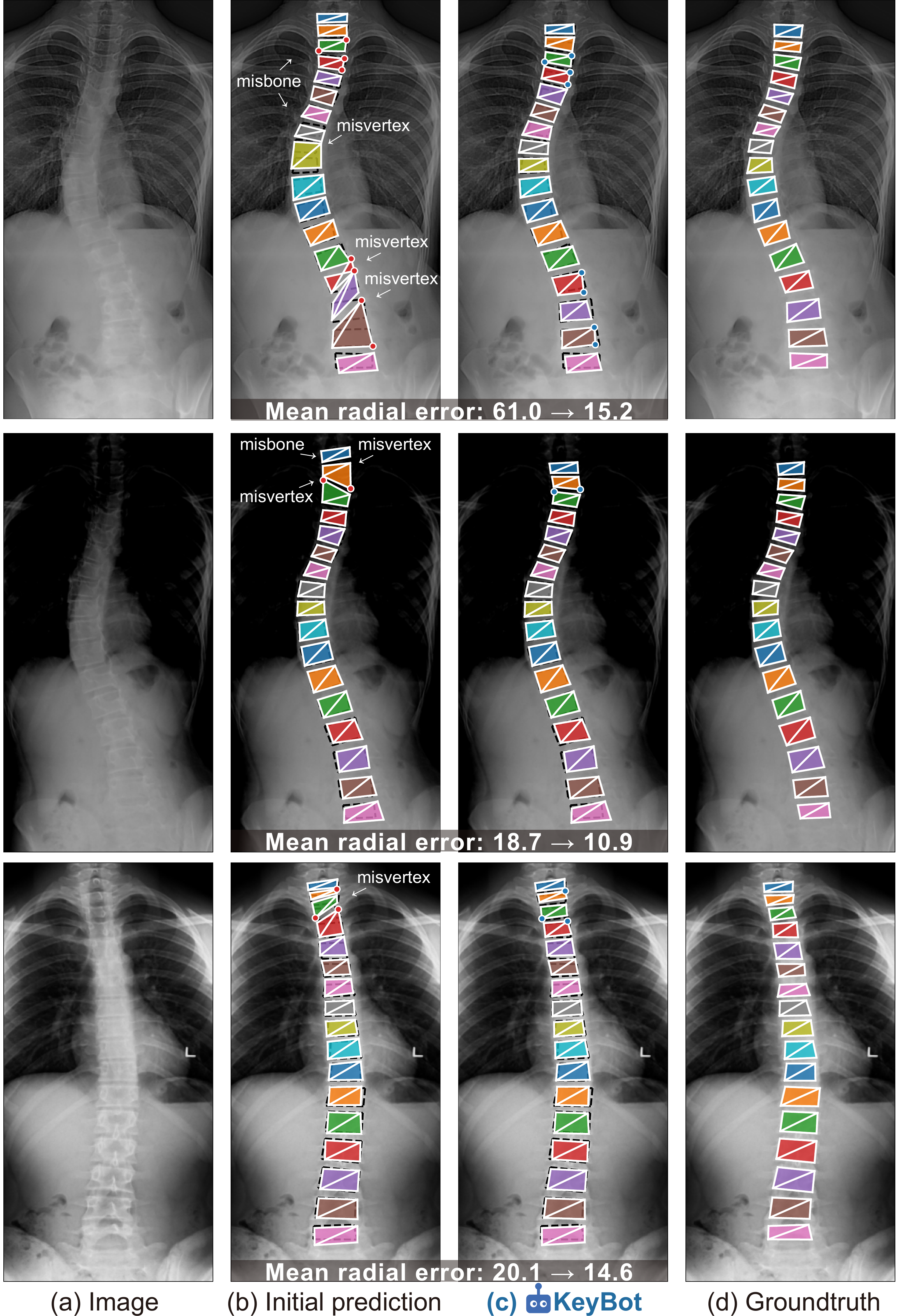}
\caption{Additional qualitative results of KeyBot on the \spinewebdata dataset, with a maximum of three iterations. } \label{supple_fig:qual2}
\end{figure}

\begin{figure}[t!]
\includegraphics[width=\textwidth]{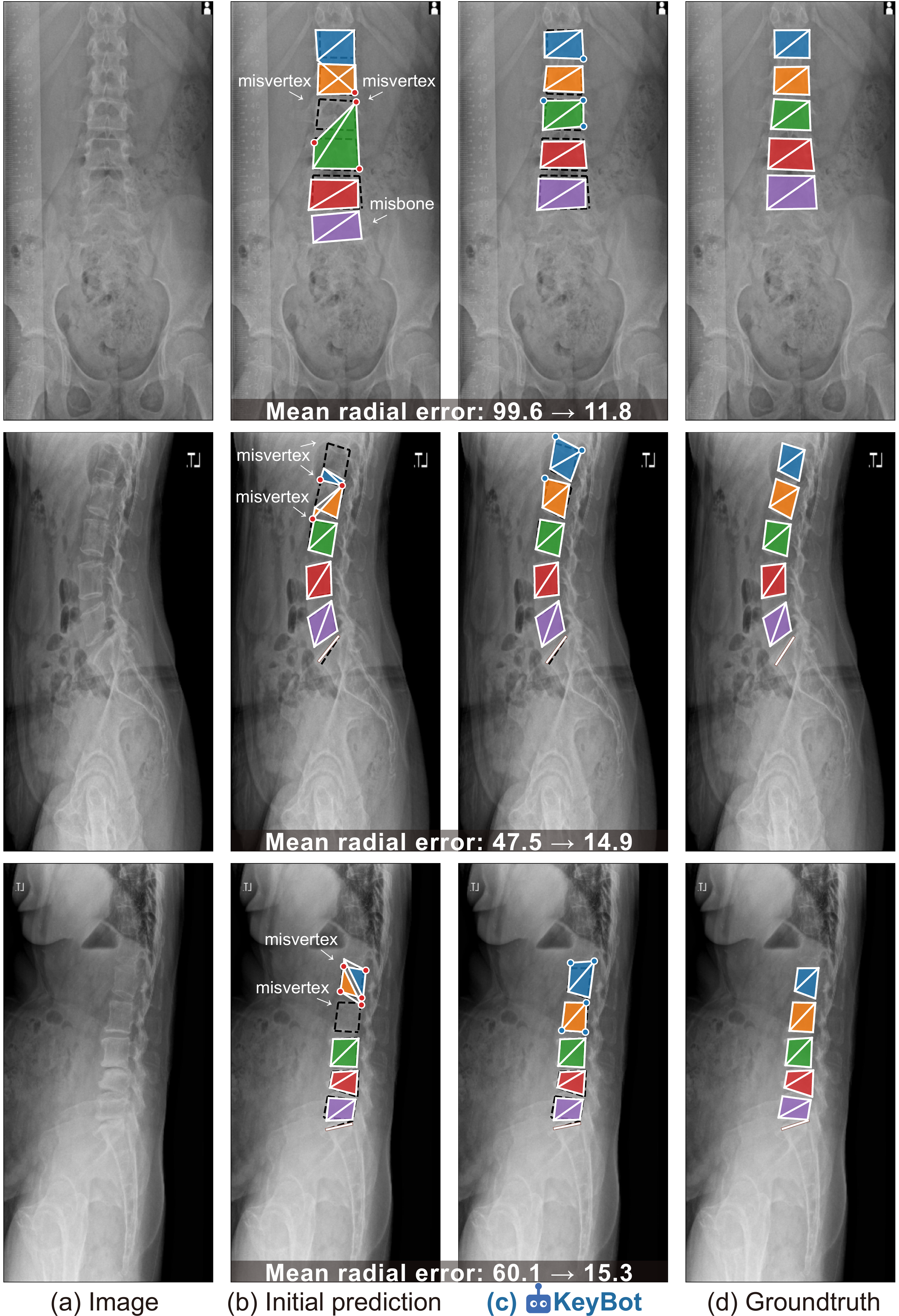}
\caption{Additional qualitative results of KeyBot on the \buudataAP (top) and \buudataLA (middle and bottom) datasets, with a maximum of three iterations.} \label{supple_fig:qual3}
\end{figure}

This section presents additional qualitative results on the AASCE dataset, as depicted in Figs.~\ref{supple_fig:qual1}~and~\ref{supple_fig:qual2}, and on the BUU-AP and BUU-LA datsets, as shown in Fig.~\ref{supple_fig:qual3}, highlighting the effectiveness of KeyBot in autonomously revising multiple keypoints simultaneously, minimizing the need for user intervention. A diagonal line connecting the upper-right vertex to the lower-left vertex is indicated in each vertebra.

Initial model predictions often exhibit significant inaccuracies in representing bone shape, posing challenges for users in identifying and correcting errors, mainly when keypoints are densely clustered or overlap.
KeyBot significantly streamlines this error correction process by effectively identifying erroneous keypoints, which are marked with red dots in the initial predictions. It distinguishes well-represented bone shape keypoints from severe morphological distortions, including misvertex errors (incorrect positioning of a portion of vertebra keypoints) and lr-inversion errors (incorrect left-right orientation).
KeyBot is also highly effective in recognizing misbone errors, where an entire bone is misidentified, as illustrated in Figs.~\ref{supple_fig:qual1} and \ref{supple_fig:qual2}. 

These capabilities allow KeyBot to provide targeted interventions for specific errors, greatly enhancing the accuracy and reliability of the overall keypoint estimation process. 
By automating the identification and correction of such errors, KeyBot not only improves the precision of the model but also reduces the burden on users,  minimizing the cognitive load on users, allowing them to concentrate on verifying and fine-tuning the results, requiring considerably less effort.

For a more comprehensive understanding of KeyBot's capabilities, please refer to our demo video, which visually represents KeyBot's efficiency and accuracy across various examples.

\end{document}